\pdfoutput=1
\documentclass[11pt]{article}

\usepackage{ACL2023}

\usepackage{times}
\usepackage{latexsym}

\usepackage[T1]{fontenc}
\usepackage[utf8]{inputenc}

\usepackage{microtype}

\usepackage{inconsolata}

\usepackage{graphicx}

\usepackage{kotex}

\usepackage{amsmath}
\usepackage{multirow}
\usepackage{booktabs}
\usepackage{float} % preamble에 추가

\usepackage{comment}
\usepackage{pifont} % symbols and shapes

\newcommand{\trex}{\textsc{TReX}}

\usepackage[many]{tcolorbox}
\tcbuselibrary{breakable}
\tcbset{
  mybox/.style={
    colback=gray!5,
    colframe=black!60,
    boxrule=0.5pt,
    arc=2mm,
    outer arc=2mm,
    left=6pt,
    right=6pt,
    top=6pt,
    bottom=6pt,
    breakable
  }
}

\usepackage{graphicx}
\usepackage{wrapfig}
\usepackage{subcaption}

\usepackage{enumitem}
\usepackage[utf8]{inputenc}

\usepackage{afterpage}
\usepackage[table]{xcolor}

\title{\includegraphics[height=1em]{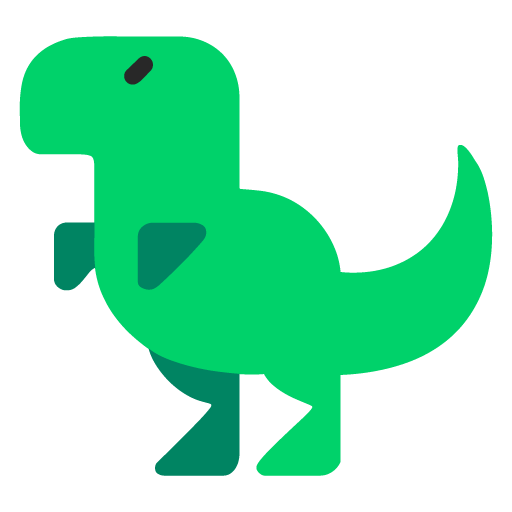}\trex{}: Tokenizer Regression for Optimal Data Mixture}

\newcommand\CoauthorMark{\footnotemark[\value{footnote}]}
\newcommand\CorrespondingAuthorMark{\footnotemark[\value{footnote}]}

\author{
    Inho Won$^1$\thanks{~~~Equal Contribution}\hspace{2.5mm}
    Hangyeol Yoo$^2$\protect\CoauthorMark\hspace{2.5mm} \\
    \bf Minkyung Cho$^1$\hspace{2.5mm}
    Jungyeul Park$^1$$^,$$^3$\hspace{2.5mm}
    Hoyun Song$^4$\thanks{~~~Corresponding Author}\hspace{2.5mm}
    KyungTae Lim$^1$$^,$$^4$\protect\CorrespondingAuthorMark \\
    $^1$KAIST CT \hspace{0.5mm}
    $^2$Seoul National University of Science and Technology \\
    $^3$Upstage AI \hspace{0.5mm}
    $^4$KAIST InnoCORE PRISM-AI Center\\
    \texttt{inho.won@kaist.ac.kr}, \texttt{hgyoo@seoultech.ac.kr}, \\
    \texttt{\{minkyung.cho, jungyeul, hysong, ktlim\}@kaist.ac.kr}
}

\begin{document}
\maketitle

\begin{abstract}
Building effective tokenizers for multilingual Large Language Models (LLMs) requires careful control over language-specific data mixtures. While a tokenizer's compression performance critically affects the efficiency of LLM training and inference, existing approaches rely on heuristics or costly large-scale searches to determine optimal language ratios. We introduce \textbf{T}okenizer \textbf{Re}gression for Optimal Data Mi\textbf{X}ture (\trex{}), a regression-based framework that efficiently predicts the optimal data mixture for tokenizer training. \trex{} trains small-scale proxy tokenizers on random mixtures, gathers their compression statistics, and learns to predict compression performance from data mixtures. This learned model enables scalable mixture search before large-scale tokenizer training, mitigating the accuracy-cost trade-off in multilingual tokenizer design.
Tokenizers trained with TReX's predicted mixtures outperform mixtures based on LLaMA3 and uniform distributions by up to 12\% in both in- and out-of-distribution compression efficiency, demonstrating strong scalability, robustness, and practical effectiveness.
All experiments are reproducible using the code available at the GitHub repository\footnote{\url{https://github.com/HanGyeol-Yoo/TReX}}.

\end{abstract}
\section{Introduction}\label{sec:introduction}

A tokenizer converts raw text into a sequence of tokens, and its performance is often evaluated by its compression capability~\cite{seo2025does, goldman2024unpacking}. This ability to represent the same sentence with fewer tokens is critical for the efficient training and inference of a model~\cite{scao2022bloom, stollenwerk2023fertility, ahia2023all}.

Achieving optimal tokenizer compression is challenging due to two main constraints. First, the saturation of a tokenizer's performance beyond a certain corpus size makes indiscriminate data scaling ineffective~\cite{goldman2024unpacking, reddy2025much, zuo2025falcon}. Second, the presence of significant data imbalance, especially in multilingual settings with underrepresented languages, leads to inconsistent performance across languages and domains~\cite{dagan2024getting, abagyan2025one}. 

Data mixture optimization has emerged as a promising solution to these challenges, and its importance is particularly pronounced in multilingual settings. In this context, multilingual data distribution is as critical as vocabulary size and total corpus volume~\cite{thakur2025art, petrov2023fairness, ahia2023all}, as language ratios directly influence subword segmentation and, consequently, compression efficiency~\cite{wang-etal-2021-multi-view, pundalik2025param1bharatgen29bmodel}. Despite its importance, most prior work has relied on empirical or manually tuned mixtures~\cite{zhang-etal-2022-robust}, often derived from heuristic methods or costly searches, leaving the interplay between language ratios and compression largely unexplored. Therefore, this work addresses the question: \textit{how can we design an optimal multilingual data mixture that maximizes tokenizer compression efficiency while remaining scalable to large LLM training?}

To this end, we introduce \textbf{T}okenizer \textbf{Re}gression for Optimal Data Mi\textbf{X}ture (\trex{}), a regression-based framework that efficiently predicts the optimal data mixture for tokenizer training. Instead of using heuristics or expensive large-scale searches, \trex{} trains a small set of lightweight proxy tokenizers on randomly sampled data mixtures, collects their compression statistics, and fits a regression model to predict compression performance from the data mixture. This learned model enables fast and reliable exploration of the vast mixture space before committing to large-scale tokenizer training. Our study investigates the following research questions:
\begin{itemize}
\item \textbf{RQ1.} Can \trex{} effectively approximate an optimal multilingual data mixture for tokenizer training?
\item \textbf{RQ2.} Is the relationship between data mixture and compression consistent across different data and vocabulary scales?
\item \textbf{RQ3.} Can \trex{} maintain robust compression performance under diverse linguistic and domain-specific settings?
\end{itemize}

To answer these questions, we (1) propose a regression-based framework for predicting optimal data mixtures for tokenizer training, (2) evaluate tokenizers trained with the predicted mixtures against those based on GPT-4o~\cite{hurst2024gpt}, LLaMA3~\cite{grattafiori2024llama}, and uniform distributions, and (3) assess the scalability and generalization of \trex{} across multilingual and domain-specific datasets. Experiments show that the proposed regression model achieves a mean absolute percentage error of 1.989 and a rank correlation above 0.97, validating its predictive reliability. Tokenizers trained with the predicted optimal mixtures outperform baselines by up to 12\% in both in- and out-of-distribution compression efficiency,
demonstrating the scalability and practical effectiveness of \trex{}. The key contributions of this paper are as follows:
\begin{itemize}
\item We propose \trex{}, a regression-based framework that efficiently searches for the optimal data mixture for tokenizer training.
\item We conduct a detailed empirical comparison against various data mixture strategies.
\item We demonstrate the scalability and generalization of \trex{} across multilingual and domain-specialized settings.
\end{itemize}

\section{Related Work}\label{sec:related_work}
\paragraph{Tokenizer and compression}
A tokenizer's key performance metric is compression, which measures how efficiently a given text can be represented by counting how many tokens are needed to encode it~\cite{scao2022bloom, stollenwerk2023fertility, ali-etal-2024-tokenizer, dagan2024getting}.

In practice, improving compression is non-trivial, as it is shaped by complex interactions among corpus size, vocabulary size, and data distribution~\cite{whittington-etal-2025-tokenisation, kim2025kormokoreanopenreasoning}. Prior work has shown that merely increasing the size of the training corpus causes compression performance to saturate beyond a certain point~\citep{reddy2025much, goldman2024unpacking,zuo2025falcon}. Similarly, expanding the vocabulary size excessively fails to produce proportional gains in compression efficiency~\citep{gowda2020finding, liu2025superbpe}. Furthermore, when the training data are unevenly distributed, tokenizers tend to overfit to dominant languages or domains, achieving strong performance in some areas but suffering severe degradation in others~\cite{dagan2024getting, abagyan2025one, petrov2023fairness}.
These findings highlight that compression efficiency cannot be improved indefinitely through scale alone. Given fixed corpus and vocabulary sizes, the proportions of languages and domains in the data mixture become the key determinant of tokenizer performance.

\paragraph{Data Mixture Optimization}
The composition of training data plays a decisive role in model performance, motivating research on balancing data mixtures.
Early approaches relied on heuristic or empirical strategies—assigning mixture weights by linguistic families or corpus size~\cite{thakur2025art,hayase2024data,karthika2025multilingualtokenizationlensindian} or iteratively refining mixtures through repeated large-scale training~\cite{thakur2025art,zhang-etal-2022-robust}.
Despite their practicality, these methods are either manually crafted or prohibitively expensive, lacking a systematic means to predict optimal mixtures.
Recent model-based studies such as DoReMi~\cite{xie2024doremi}, DoGE~\cite{fan2023doge}, and RegMix~\cite{liu2025regmix} address this limitation by estimating domain-wise loss via regression or interpolation.
These approaches efficiently infer optimal mixture weights and improve pre-training or SFT performance~\cite{li2025data,ye2024data}, but remain tied to loss minimization objectives, limiting their applicability to tokenizer training, where compression provides the most direct and scalable measure of representational efficiency.
Building on model-based paradigms such as RegMix, our work extends this regression framework to the tokenizer level, predicting compression outcomes from data mixtures to identify optimal multilingual configurations without exhaustive retraining.

\section{Preliminaries}

\begin{figure*}[t]
    \centering
    \includegraphics[width=\linewidth]{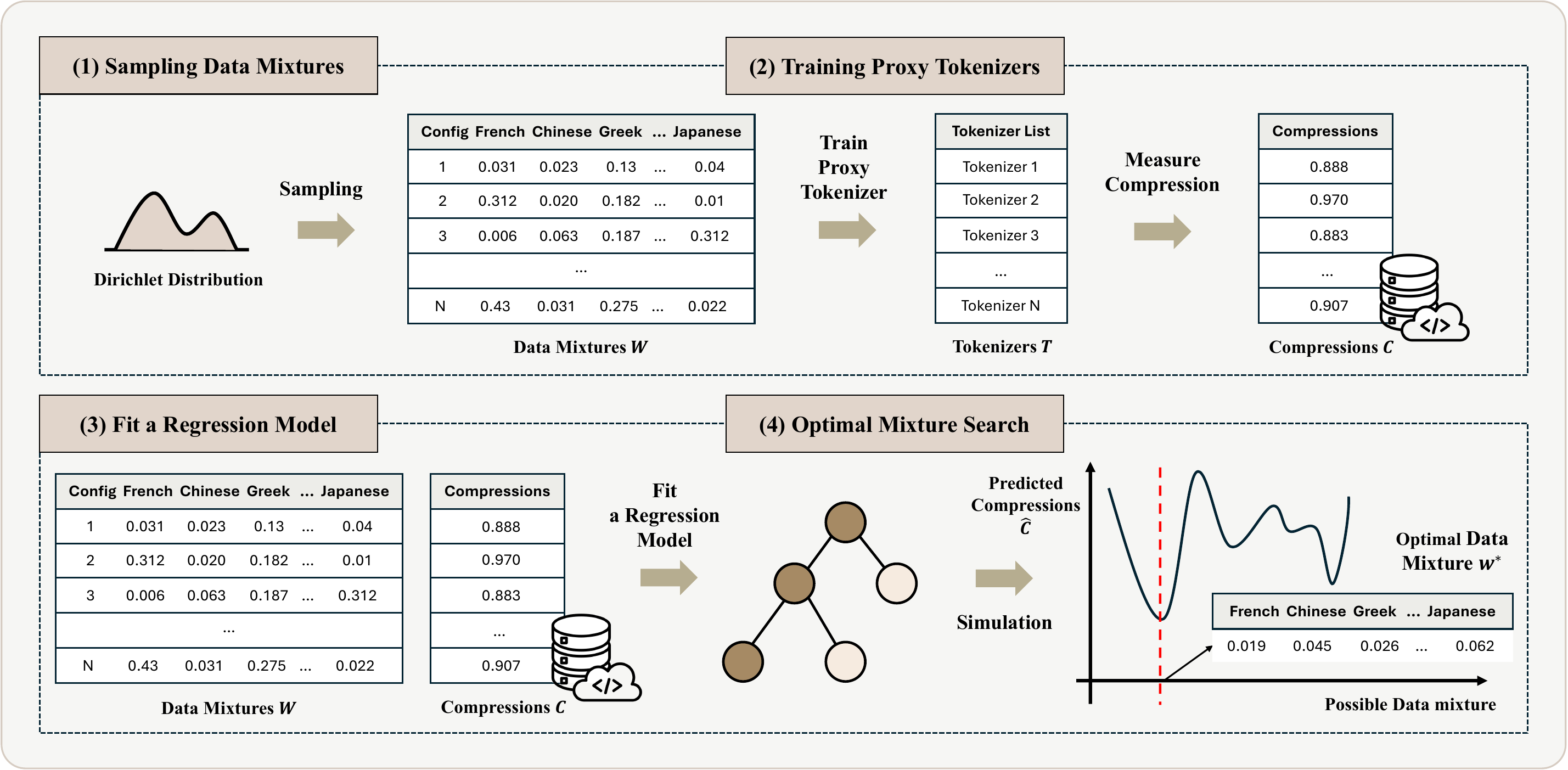}
    \caption{\trex{} overview. 
    The process consists of four stages: 
    (1) $N$ data mixtures are sampled, 
    (2) proxy tokenizers are trained for each mixture to measure $C$, 
    (3) a regression model is fitted using $\mathbf{w}$ as input and $C$ as the target, and 
    (4) the model predicts $C$ for candidate mixtures to identify the optimum.}
    \label{fig:trex-overview}
\end{figure*}

\paragraph{Data Mixture}
We define the overall corpus as $D = \{ D_{\text{train}}, D_{\text{test}} \}$,
where $D_{\text{train}}$ and $D_{\text{test}}$ denote the training and test corpora, respectively.
Each tokenizer is trained under a configuration characterized by two parameters:
the total training corpus size $S$ and the vocabulary size $V$.

The training corpus $D_{\text{train}}$ consists of $k$ language-specific corpora,
$D_{\text{train}} = \{ d_1, d_2, \dots, d_k \}$.
A \textit{data mixture} $\mathbf{w}$ specifies the relative contribution of each language corpus:
\begin{equation}
    \mathbf{w} = (w_1, \ldots, w_k),\;\sum_{i=1}^{k} w_i = 1, \; w_i \ge 0
\end{equation}
The set of all valid mixtures defines the \textit{mixture space} $\mathcal{W}$:
\begin{equation}
    \mathcal{W} = \big\{ \mathbf{w} \mid \sum_{i=1}^{k} w_i = 1, \; w_i \ge 0 \text{ for all } i \big\}.
\end{equation}
By fixing $(S, V)$ and varying $\mathbf{w} \in \mathcal{W}$, 
we can train a family of tokenizers, denoted $T_{\mathbf{w}}$, 
each reflecting a different multilingual data mixture.

\paragraph{Tokenizer Compression}
We evaluate tokenizer performance using the \textit{Normalized Sequence Length (NSL)}~\citep{dagan2024getting},
which measures the compression $C_{\text{tar}}$ of a target tokenizer $T_{\text{tar}}$ 
relative to a reference tokenizer $T_{\text{ref}}$ on a test corpus $D_{\text{test}}$:
\begin{equation}
    C_{\text{tar}} =
    \frac{\sum_{i=1}^{N} \text{Len}(T_{\text{tar}}(D_{\text{test}}^{i}))}
         {\sum_{i=1}^{N} \text{Len}(T_{\text{ref}}(D_{\text{test}}^{i}))}.
\end{equation}
Here, $T(\cdot)$ denotes the tokenization function and $\text{Len}(\cdot)$ 
the number of tokens in the $i$-th sample.
A smaller $C_{\text{tar}}$ indicates better compression; 
values below $1.0$ imply that the target tokenizer achieves more compact representations than the reference.

\paragraph{Problem Statement}
Our objective is to find the optimal data mixture $\mathbf{w}^\ast$ 
that minimizes the compression score for a given corpus size $S$ and vocabulary size $V$:
\begin{equation}
    C_{\mathbf{w}}(S, V) =
    \frac{\sum_{i=1}^{N} \text{Len}(T_{\mathbf{w}}(D_{\text{test}}^{i}))}
         {\sum_{i=1}^{N} \text{Len}(T_{\text{ref}}(D_{\text{test}}^{i}))}.
\label{eq:objective_func}
\end{equation}
\begin{equation}
    \mathbf{w}^\ast = \arg\min_{\mathbf{w} \in \mathcal{W}} C_{\mathbf{w}}(S, V).
\label{eq:optimization}
\end{equation}
Here, $T_{\mathbf{w}}$ denotes the tokenizer trained with mixture $\mathbf{w}$.
This formulation captures how the tokenizer's compression varies with the data mixture 
and formalizes the goal of discovering the optimal mixture $\mathbf{w}^\ast$.

\section{Method}\label{sec:method}
We propose \trex{} (\textbf{T}okenizer \textbf{Re}gression for e\textbf{X}ploration), a framework that predicts the optimal data mixture for large-scale tokenizer training by leveraging numerous small-scale proxy tokenizers. \trex{} consists of four steps as detailed in \autoref{fig:trex-overview}.

\subsection{Sampling Data Mixtures}
First, we sample $N$ data mixtures, $\textstyle \mathbf{W}=\{\mathbf{w}_1, \cdots, \mathbf{w}_n\}$, from the mixture space $\mathcal{W}$ using a Dirichlet distribution based on the data size of each language. The Dirichlet distribution is ideal for this task as it generates mixture ratios on a probability simplex (i.e., vectors whose elements sum to 1) while reflecting the actual data distribution across languages~\cite{lin2016dirichlet, tsun2020probability}. This ensures a diverse yet feasible set of configurations for training the regression model.

\subsection{Training Proxy Tokenizers}
Each sampled mixture $\mathbf{w_i} \in \mathbf{W}$ is used to train a corresponding proxy tokenizer $T_{\mathbf{w_i}}$ in a small-scale setting $(S_s, V_s)$, where the subscript $s$ denotes small-scale configurations with reduced corpus size $S_s$ and vocabulary size $V_s$. We then use each proxy tokenizer to measure its compression score $C_{\mathbf{w_i}}$ on a target test corpus $D_{\text{test}}$. The resulting set $\mathbf{C}=\{C_{\mathbf{w_1}}, \cdots, C_{\mathbf{w_n}}\}$, forms the target values for training the regression model.

\subsection{Fitting a Regression Model}

We fit a regression model $f$ on the collected training set 
$\{ (\mathbf{w_i}, C_{\mathbf{w_i}}) \}_{\mathbf{i}=1}^{N}$, 
where each $\mathbf{w_i}$ denotes a sampled data mixture 
and $C_{\mathbf{w}_i}$ is its measured compression score in the small-scale setting.
The model learns the mapping
\begin{equation}
    f: \mathbf{w} \mapsto C_{\mathbf{w}},
\end{equation}
which approximates the relationship between data mixture and compression performance.
Once trained, $f$ can predict the expected compression score for any new mixture $\mathbf{w}'$ 
without requiring additional tokenizer training, 
enabling efficient exploration of the mixture space.

\subsection{Optimal Mixture Search}
The trained model $f$ enables an efficient, large-scale search across the entire mixture space.
For example, it can estimate compression scores for over 50M data mixtures, and this process can be completed within a few seconds using minimal computation.
We identify the optimal data mixture $\mathbf{w}^\ast$ by finding the minimum of the learned function:
\begin{equation}
\mathbf{w}^\ast = \arg\min_{\mathbf{w}}f(\mathbf{w})
\label{eq:w-ast}
\end{equation}

Finally, we use this optimal mixture $\mathbf{w}^\ast$ to train a single, large-scale tokenizer.

\section{Experiments and Results}
In this section, we evaluate the effectiveness of \trex{} by addressing the following three research questions introduced earlier:
(RQ1) Can \trex{} effectively approximate an optimal multilingual data mixture for tokenizer training?
(RQ2) Is the relationship between data mixture ratios and compression performance consistent across different corpus scales and vocabulary sizes?
(RQ3) Is  efficient and scalable in real-world multilingual environment?

\subsection{Experimental Setup}
The experiments were organized from three perspectives:
(1) dataset configuration, (2) regression model design, and (3) evaluation methodology.

\paragraph{Datasets}
For multilingual tokenizer training, we used FineWeb2-HQ~\cite{messmer2025multilingdatacomp}, one of the most widely used large-scale multilingual corpora.
It comprises 19 languages and provides a diverse, publicly available dataset suitable for studying data mixture effects.
We held out 0.1\% of the corpus as the In-Distribution (ID) test set to avoid data leakage, and used FLORES~\cite{nllbteam2022languageleftbehindscaling} as the Out-of-Distribution (OOD) test set.
A detailed description of the corpus composition is provided in Appendix~\ref{sec:appendix_datasets}.

\paragraph{Regression Model}
Following prior work on data mixture optimization~\cite{liu2025regmix}, we employed LightGBM~\cite{ke2017lightgbm} for the regression task.
LightGBM, a gradient boosting based ensemble of decision trees, efficiently captures non-linear relationships between language mixture ratios and compression performance, making it well-suited for this experiment.

\paragraph{Training Details}
As described in Section~\ref{sec:method} and illustrated in \autoref{fig:trex-overview}, \trex{} was trained through four main stages.
First, we sampled $N{=}512$ data mixtures $\mathbf{W} = \{\mathbf{w_1}, \mathbf{w_2}, \dots, \mathbf{w_{512}}\}$ from a Dirichlet distribution.
Second, for each sampled mixture $\mathbf{w_i}$, we trained a proxy tokenizer $T_{\mathbf{w_i}}$ and measured its corresponding compression $C_{\mathbf{w_i}}$, forming the set $\mathbf{C} = \{C_{\mathbf{w_1}}, \dots, C_{\mathbf{w_{512}}}\}$.
To train lightweight proxy tokenizers, we randomly extracted $S{=}1$ GB of data from the $D_{\text{train}}$ and set the vocabulary size to $V{=}64$K.
Each subset was automatically partitioned according to the language ratios specified by $\mathbf{w_i}$.
This process yielded 512 pairs of language mixtures and their corresponding compression $(\mathbf{w_i}, C_{\mathbf{w_i}})$, of which 480 were used for training and 32 for evaluation.
For large-scale experiments, following prior studies~\cite{bi2024deepseek, liu2025superbpe}, we increased the configuration to $S{=}30$ GB and $V{=}200$K.
Compression was measured relative to the GPT-4o tokenizer, which serves as the reference tokenizer $T_{\text{ref}}$, as recent analyses have shown that it provides more balanced multilingual coverage than other widely used tokenizers, such as LLaMA3~\cite{hayase2024data}.

\paragraph{Evaluation metrics}
We evaluated the predictive performance of the regression model using the Mean Absolute Percentage Error (MAPE) and the Spearman rank correlation ($\rho$). MAPE quantifies the absolute deviation between the predicted and actual compression values, while $\rho$ measures the consistency of their relative rankings. MAPE is used to validate RQ1, which evaluates the model's ability to approximate optimal compression of data mixture. Meanwhile, $\rho$ is used to validate RQ2 by examining rank stability across different corpus and vocabulary scales~\cite{liu2025regmix}. Together, these metrics provide a comprehensive view of the predictive accuracy and rank consistency of \trex{}.

\subsection{Regression Model Performance}

\paragraph{Within-Scale Prediction}
\begin{table}
\small
\centering
\begin{tabular}{llcc}
\toprule
\multicolumn{2}{c}{\textbf{Test Setting}} & \multirow{2}{*}{\textbf{Correlation} ($\rho$)} & \multirow{2}{*}{\textbf{MAPE}} \\
\cmidrule(lr){1-2} 
Data Size & Vocabulary & $\uparrow$ & $\downarrow$ \\
\midrule
1GB   & 64k    &  0.979  & 1.989  \\
% 5GB   & 64k    &  0.970  & -  \\
% 10GB  & 100k   &  0.967  & -  \\
% 30GB  & 200k   &  0.960  & -  \\
\bottomrule
\end{tabular}
% \caption{회귀 모델의 예측 성능. $(S=1\text{GB}, V=64\text{k})$ 설정에서 32개의 테스트 데이터 혼합으로 학습한 토크나이저의 실제 압축률과 회귀 모델이 예측한 압축률을 비교하였다. 실험결과, 회귀모델은 새로운 data mixture를 봐도 실제 compression을 정확하게 예측할 수 잇고, rank correlation도 높음을 확인하다.}
\caption{Prediction performance of the regression model under $(S=1\text{GB}, V=64\text{k})$. The model accurately predicts actual compression for unseen data mixtures, achieving a high rank correlation.}
\label{tab:regression_results}
\vspace{-0.2in}
\end{table}
\autoref{tab:regression_results} presents the performance of the \trex{} regression model under the $S{=}1$GB, $V{=}64$K setting.
Across 32 test samples, \trex{} achieved a MAPE of 1.989, indicating that it predicts compression score with less than 2\% average error.
Furthermore, the model also achieved a Spearman rank correlation of $\rho{=}0.979$, showing that it preserves the relative ranking among data mixtures with nearly 98\% consistency.
These results show that \trex{} can accurately model the relationship between data mixtures and compression even in a small proxy environment, providing strong empirical support for RQ1.

\paragraph{Cross-Scale Generalization}
Previously, we showed that \trex{} accurately predicts the compression of various data mixtures under a fixed configuration ($S{=}1$GB, $V{=}64$K).
However, in practical tokenizer design, both corpus size ($S$) and vocabulary size ($V$) can vary substantially.
A key question, therefore, is whether the relative ranking of mixtures in terms of compression remains consistent across different scales.
If this stability holds, \trex{} can serve as a scalable approach, enabling small proxy tokenizers to predict performance in large-scale settings.
This assumption corresponds to the principle of \textit{Rank Invariance} proposed by \citet{liu2025regmix}.

Rank Invariance assumes that although absolute compression values may change as environmental variables (e.g., corpus size or vocabulary size) vary, the relative ordering of mixtures remains stable.
To test this assumption, we trained tokenizers under multiple scale settings with varying corpus and vocabulary sizes (1GB–64K, 5GB–64K, 10GB–100K, 30GB–200K) and computed the Spearman rank correlation of mixture efficiencies across these configurations.
\begin{figure}
\includegraphics[width=\linewidth]{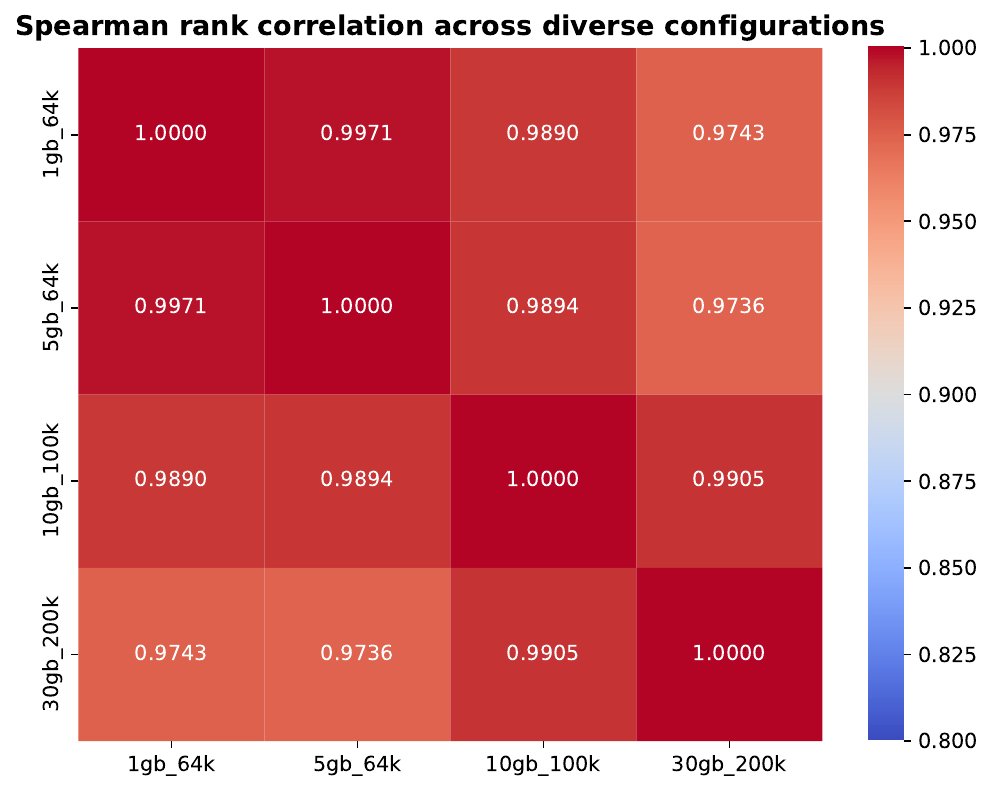}
\caption{Spearman rank correlation heatmap across diverse configurations $(S, V)$.}
\label{fig:spearman_corr_heatmap}
\vspace{-0.2in}
\end{figure}
As shown in \autoref{fig:spearman_corr_heatmap}, correlation between all setting pairs were consistently high ($\rho \geq 0.96$).
For example, the correlation between the smallest (1GB–64K) and largest (30GB–200K) configurations reached $\rho{=}0.974$, indicating that compression rankings remain nearly unchanged even when corpus size and vocabulary scale differ by several orders of magnitude.
This finding demonstrates that mixture efficiency is largely invariant to training scale--i.e., Rank Invariance holds.

These findings confirm that \trex{} can reliably generalize language importance patterns observed in small proxy tokenizers to large-scale environments.
By maintaining consistent mixture prediction performance independent of scale variations, \trex{} provides strong evidence for the ``scale-invariant mixture consistency'' proposed in RQ2.
Further experiments supporting Rank Invariance are provided in Appendix~\ref{sec:appendix_rank_invariance}.

\begin{table*}[tb]
\centering
\tiny
\begin{tabular}{llrrrrr}
\toprule
\textbf{\textsc{Lang}} & \textbf{\textsc{Char}} & \textbf{\textsc{$\mathbf{w}^{uni}$}} & \textbf{\textsc{$\mathbf{w}^{LB}$}} & \textbf{\textsc{$\mathbf{w}^{gpt}$}} & \textbf{\textsc{$\mathbf{w}^{llama}$}} & $\mathbf{w}^{\trex{}}$ \\
\midrule
\rowcolor{gray!15} 
\multicolumn{7}{c}{\textbf{In-Distribution (over FineWeb2-HQ)} ($\downarrow$)}  \\
\midrule
\textbf{CMN}  & Hani & 0.827 & 1.097 & 0.841 & 0.962 & 0.831 \\
\textbf{ELL}  & Grek & 0.721 & 0.847 & 0.763 & 0.715 & 0.778 \\
\textbf{FAS}  & Arab & 0.842 & 1.021 & 0.900 & 0.825 & 0.906 \\
\textbf{JPN}  & Jpan & 0.682 & 0.997 & 0.748 & 0.621 & 0.663 \\
\textbf{RUS}  & Cyrl & 0.982 & 1.121 & 0.868 & 0.901 & 0.878 \\
\textbf{DEU}  & Latn & 0.971 & 0.932 & 0.928 & 0.953 & 0.933 \\
\textbf{FRA}  & Latn & 0.995 & 0.962 & 0.939 & 0.991 & 0.933 \\
\textbf{ITA}  & Latn & 0.886 & 0.868 & 0.912 & 0.912 & 0.862 \\
\textbf{POR}  & Latn & 0.981 & 0.964 & 0.938 & 0.988 & 0.962 \\
\textbf{SPA}  & Latn & 0.992 & 0.971 & 0.943 & 0.985 & 0.978 \\
\textbf{CES}  & Latn & 0.746 & 0.715 & 0.838 & 0.713 & 0.813 \\
\textbf{DAN}  & Latn & 0.827 & 0.812 & 0.872 & 0.896 & 0.859 \\
\textbf{HUN}  & Latn & 0.693 & 0.667 & 1.204 & 1.204 & 0.918 \\
\textbf{IND}  & Latn & 0.842 & 0.821 & 0.891 & 0.957 & 0.890 \\
\textbf{NLD}  & Latn & 0.970 & 0.952 & 0.930 & 1.006 & 0.970 \\
\textbf{POL}  & Latn & 0.758 & 0.743 & 0.809 & 0.762 & 0.778 \\
\textbf{SWE}  & Latn & 0.841 & 0.827 & 0.906 & 0.915 & 0.864 \\
\textbf{TUR}  & Latn & 0.782 & 0.765 & 0.815 & 0.737 & 0.894 \\
\textbf{VIE}  & Latn & 0.894 & 0.890 & 0.905 & 0.893 & 0.896 \\
\midrule
\multicolumn{2}{c}{\textbf{All Languages}} & 0.888 & 0.970 & 0.883 & 0.907 & \textbf{0.871} \\
\bottomrule
\end{tabular}
\hspace{5mm}
\begin{tabular}{llrrrrr}
\toprule
\textbf{\textsc{Lang}} & \textbf{\textsc{Char}} & \textbf{\textsc{$\mathbf{w}^{uni}$}} & \textbf{\textsc{$\mathbf{w}^{LB}$}} & \textbf{\textsc{$\mathbf{w}^{gpt}$}} & \textbf{\textsc{$\mathbf{w}^{llama}$}} & $\mathbf{w}^{\trex{}}$ \\
\midrule
\rowcolor{gray!15} 
\multicolumn{7}{c}{\textbf{Out-Of-Distribution (over FLORES-200)} ($\downarrow$)} \\
\midrule
\textbf{CMN}  & Hani & 0.838 & 1.134 & 0.858 & 0.951 & 0.838 \\
\textbf{ELL}  & Grek & 0.707 & 0.839 & 0.752 & 0.701 & 0.768 \\
\textbf{FAS}  & Arab & 0.940 & 1.127 & 1.003 & 0.923 & 1.009 \\
\textbf{JPN}  & Jpan & 0.672 & 1.004 & 0.746 & 0.608 & 0.652 \\
\textbf{RUS}  & Cyrl & 0.993 & 1.141 & 0.870 & 0.905 & 0.879 \\
\textbf{DEU}  & Latn & 0.981 & 0.937 & 0.933 & 0.960 & 0.937 \\
\textbf{FRA}  & Latn & 1.007 & 0.972 & 0.946 & 1.002 & 0.939 \\
\textbf{ITA}  & Latn & 0.883 & 0.864 & 0.915 & 0.913 & 0.859 \\
\textbf{POR}  & Latn & 0.992 & 0.973 & 0.945 & 1.000 & 0.970 \\
\textbf{SPA}  & Latn & 1.005 & 0.985 & 0.954 & 0.999 & 0.992 \\
\textbf{CES}  & Latn & 0.752 & 0.720 & 0.852 & 0.719 & 0.824 \\
\textbf{DAN}  & Latn & 0.871 & 0.856 & 0.919 & 0.944 & 0.902 \\
\textbf{HUN}  & Latn & 0.696 & 0.669 & 1.223 & 1.225 & 0.929 \\
\textbf{IND}  & Latn & 0.854 & 0.832 & 0.907 & 0.980 & 0.905 \\
\textbf{NLD}  & Latn & 0.985 & 0.968 & 0.943 & 1.027 & 0.987 \\
\textbf{POL}  & Latn & 0.757 & 0.743 & 0.812 & 0.763 & 0.779 \\
\textbf{SWE}  & Latn & 0.858 & 0.843 & 0.930 & 0.938 & 0.884 \\
\textbf{TUR}  & Latn & 0.794 & 0.779 & 0.827 & 0.748 & 0.908 \\
\textbf{VIE}  & Latn & 0.916 & 0.912 & 0.926 & 0.916 & 0.916 \\
\midrule
\multicolumn{2}{c}{\textbf{All Languages}} & 0.904 & 0.997 & 0.907 & 0.906 & \textbf{0.877} \\
\bottomrule
\end{tabular}
% \caption{왼쪽은 In-Distribution (FineWeb2-HQ), 오른쪽은 Out-of-Distribution (FLORES) 데이터셋에서의 다양한 data mixture별 compression performance을 비교한 결과이다. 각 행은 언어(Lang)와 그에 대응하는 문자 체계(Char)를 나타내며, 제안한 최적 혼합 $\mathbf{w}^\ast$가 두 설정 모두에서 평균적으로 가장 우수한 압축 성능을 보인다.}
\caption{A comparison of $C$ (lower is better) for various data mixtures on the In-Distribution (FineWeb2-HQ) dataset and the Out-of-Distribution (FLORES) dataset. Each row represents a language (Lang) and its corresponding character system (Char). The proposed optimal mixture, $\mathbf{w}^{\trex{}}$, achieves the best performance across both settings.}
\label{tab:comp-main-results}
\end{table*}

\begin{figure*}
    \centering
    \includegraphics[width=\linewidth]{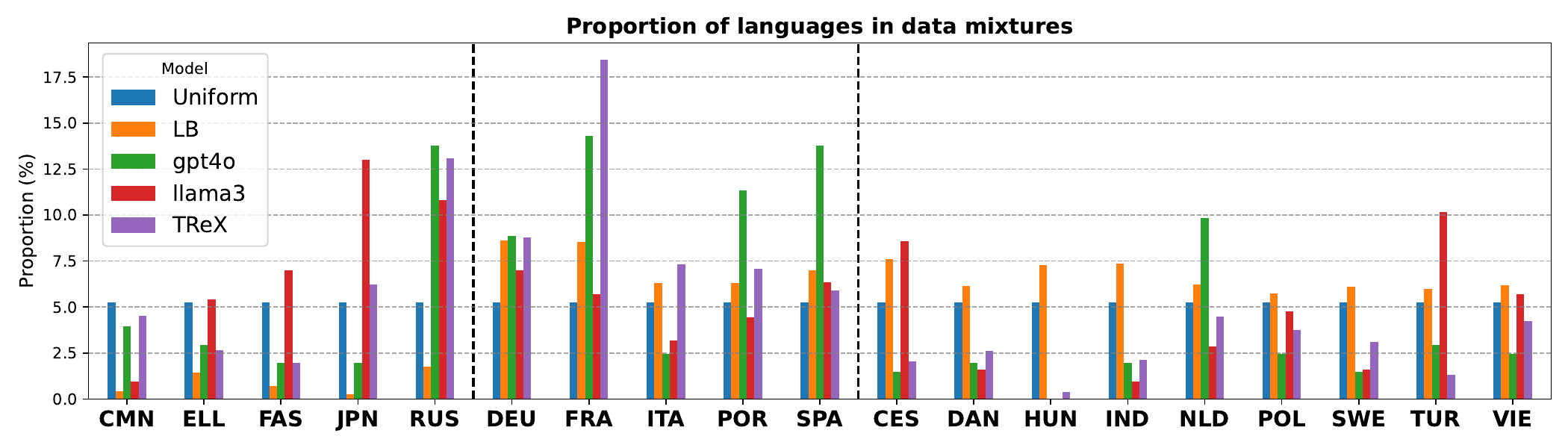}
    \vspace{-0.3in}
    \caption{Proportions of languages in the data mixtures.}
    \label{fig:language_distribution}
\end{figure*}

\subsection{Evaluating \trex{} on Large-Scale Tokenizer Training} 
In this section, we compare how the optimal data mixture $\mathbf{w}^{\trex{}}$ derived from the \trex{} performs in large-scale tokenizer training ($S{=}30$GB, $V{=}200$K) relative to other baseline mixtures in terms of the compression performance $C$. To this end, we established the following four mixtures as baselines:

\begin{itemize}
    \item \textsc{$\mathbf{w}^{uni}$}: A data mixture that assigns equal weights to all languages, i.e. $\mathbf{w}^{uni}=(\frac{1}{19}, \frac{1}{19}, \dots, \frac{1}{19})$
    \item \textsc{$\mathbf{w}^{LB}$}: A data mixture reflecting the language proportions proposed by \citeauthor{abagyan2025one}. It specifically applies a strategy where languages are grouped into language buckets based on family and script to determine the optimal mixture ratios for training.
    \item \textsc{$\mathbf{w}^{gpt}$}: A data mixture reflecting the proportions of 19 languages in the vocabulary of the GPT-4o tokenizer
    \item \textsc{$\mathbf{w}^{llama}$}: A data mixture reflecting the proportions of 19 languages in the vocabulary of the LLaMA3 tokenizer
\end{itemize}
We trained large-scale tokenizers using these four mixtures, and the exact language ratios for each model are presented in Appendix~\ref{sec:appendix_data_mixture_dist}.
The average compression of each tokenizer was evaluated as a weighted mean, where the weights correspond to the language proportions in the test corpus $D_{\text{test}}$ and each term represents the compression performance for that language. 

\paragraph{Overall}
\autoref{tab:comp-main-results} compares the compression performance of the tokenizer trained with the optimal data mixture $\mathbf{w}^\trex{}$ derived from \trex{}'s regression model against four baseline methods.
In the In-Distribution setting, $\mathbf{w}^\trex{}$ achieved a score of 0.871, slightly outperforming $\mathbf{w}^{gpt}$, which recorded 0.883, and showing a 10.21\% improvement over $\mathbf{w}^{LB}$.
In the Out-of-Distribution setting, $\mathbf{w}^\trex{}$ also achieved the best performance with a score of 0.877, surpassing all baseline methods in compression efficiency.

These results demonstrate that $\mathbf{w}^\trex{}$ yields a more efficient tokenizer in terms of compression compared to both previously proposed data mixture strategies ($\mathbf{w}^{uni}$, $\mathbf{w}^{LB}$) and widely used practical tokenizers ($\mathbf{w}^{gpt}$, $\mathbf{w}^{llama}$).

\paragraph{Out-of-Distribution Results and Generalization}
As shown in \autoref{tab:comp-main-results}, the most significant strength of $\mathbf{w}^\trex{}$ lies in its robustness on Out-of-Distribution (OOD) data.
Despite being optimized under the In-Distribution setting, $\mathbf{w}^\trex{}$ achieved the lowest average compression score of 0.877 on OOD corpus that deviate from the training distribution.
This value is substantially lower than that of the second-best method, $\mathbf{w}^{uni}$ (0.904), and shows a remarkable performance gap compared to $\mathbf{w}^{LB}$ (0.997). These results suggest that the data mixture derived through $\mathbf{w}^\trex{}$ does not overfit to the specific training data but instead captures the underlying relationship between languages and their compression, enabling the design of mixtures that generalize well.
Ultimately, tokenizers trained with $\mathbf{w}^\trex{}$ can deliver the most stable and efficient compression performance across the diverse and unpredictable data distributions encountered in real-world deployment scenarios.

\paragraph{Script-Level Analysis: Pronounced Gains in Non-Latin Languages}
Notably, when focusing on non-Latin character languages (Hani, Grek, Arab, Jpan, and Cyrl), the efficiency of \trex{} becomes even more pronounced.
When comparing the average compression scores of these five languages, $\mathbf{w}^{\trex{}}$ achieved $0.814$ in the Non-Latin group, lower than all baselines (\textsc{$\mathbf{w}^{uni}$}=0.848, \textsc{$\mathbf{w}^{llama}$}=0.863, etc.).
These results represent an improvement of approximately $3.4$ percentage points over the uniform distribution and $4.9$ percentage points over the $\mathbf{w}^{llama}$, indicating that $\mathbf{w}^{\trex{}}$ consistently preserves compression efficiency even for non-Latin scripts.
As shown in \autoref{fig:language_distribution}, the $\mathbf{w}^{llama}$ allocates 36.8\% of the total data to non-Latin character languages, whereas $\mathbf{w}^{\trex{}}$ assigns only 28.2\%, which is about 8.6 percentage points lower.
Nevertheless, the average compression score of the non-Latin group in $\mathbf{w}^{llama}$ (0.863) remains higher than that of $\mathbf{w}^{\trex{}}$ (0.814), further demonstrating \trex{}'s superior efficiency.

This finding suggests that simply increasing the proportion of certain languages is insufficient; rather, efficient weighting that accounts for the segmentation structure and statistical redundancy of each script is the key.
Consequently, $\mathbf{w}^{\trex{}}$ achieves higher efficiency with fewer non-Latin tokens, empirically demonstrating the importance of character-aware mixture optimization in multilingual tokenizer design.

\paragraph{Impact of Language Distribution on Compression Performance}
\autoref{fig:language_distribution} presents the language proportions of five related languages (from DEU to SPA) across different models, and \autoref{tab:comp-main-results} shows their corresponding compression efficiencies. When considered together, these results reveal a nonlinear relationship between the mixture entropy, representing the diversity of language ratios, and the average compression efficiency.

Mixtures with high entropy such as $\mathbf{w}^{uni}$ are generally stable, but they fail to sufficiently remove token redundancy within specific language families, resulting in a limited average compression score of $0.904$.
In contrast, $\mathbf{w}^{\trex{}}$ achieved the lowest average compression score ($0.877$) despite having a relatively lower entropy (a more biased distribution). This suggests that a mixture does not necessarily need to be uniform; rather, efficient bias can actually enhance tokenizer performance.

This observation contrasts with the results of $\mathbf{w}^{llama}$, whose distribution is closer to $\mathbf{w}^{uni}$, yet exhibits higher compression scores (lower efficiency) in major Romance languages such as Italian (ITA), Portuguese (POR), and Spanish (SPA).
In other words, a biased mixture that accounts for structural redundancy among languages can yield a more efficient tokenizer than a simple uniform distribution.

As further shown in Appendix \autoref{fig:entropy_compression} and \autoref{tab:mixture_entropy},
$\mathbf{w}^{\trex{}}$ achieves the highest efficiency at a moderate level of entropy, supporting the notion that efficient bias, rather than uniformity, leads to the optimal design of multilingual mixtures.

\section{Analysis in Real-World Scenarios} \label{sec:analysis}
In the previous section, we demonstrated that \trex{} achieves superior compression compared to existing tokenizers.
In this section, we examine whether this advantage translates into consistent efficiency during large-scale language model (LLM) training. In addition, we investigate its robustness in domain-specialized environments, as formulated in RQ3.

\begin{figure}[t]
    \centering
    \includegraphics[width=\columnwidth]{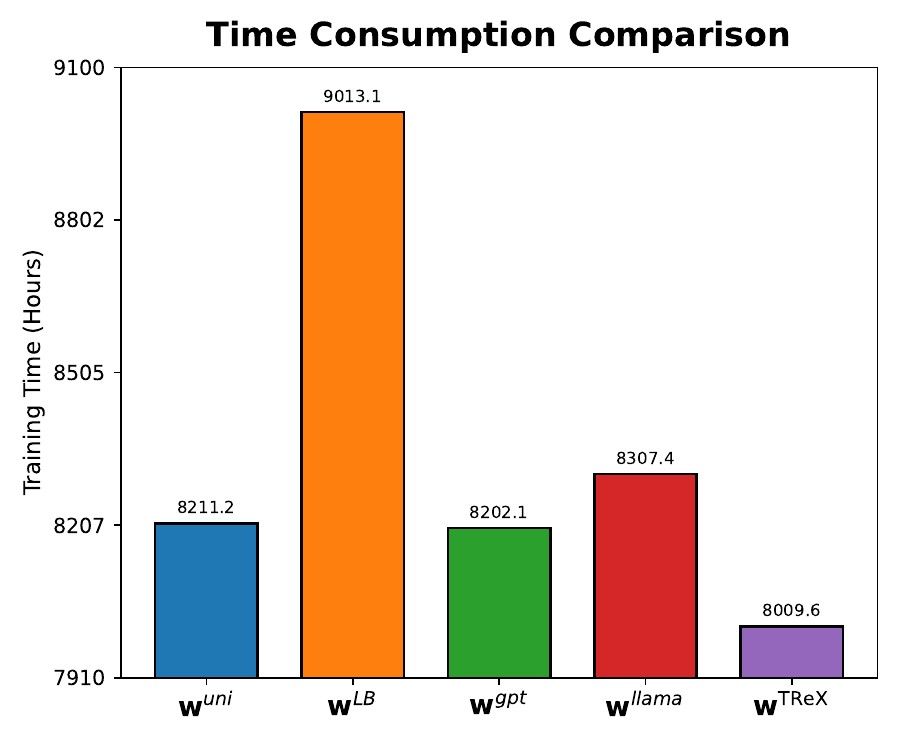}
    \caption{Model Training Time Consumption Comparison}
    \label{fig:model_time_consumption}
\end{figure}

\paragraph{Does \trex{} Improve Efficiency in LLM Training?}
To quantify the impact of tokenizer choice on LLM training costs, we follow the experimental setup of prior work~\cite{olmo20242}, and consider a 13-billion-parameter model trained on a 3-trillion-token corpus, with token counts computed under the GPT-4o tokenizer. Under this setting, a tokenizer with higher compression performance can encode the same raw corpus into fewer tokens, thereby proportionally reducing the total number of training FLOPs.

\autoref{fig:model_time_consumption} compares the estimated total training time for models trained with different tokenizer mixtures.
The tokenizer trained with $\mathbf{w^{\trex{}}}$ achieves the shortest training time of 8,009.6 hours, outperforming all baselines.
Relative to the least efficient mixture ($\mathbf{w}^{LB}$), \trex{} reduces total training time by more than 1,000 hours, and still saves roughly 200 hours compared to the next-best configuration ($\mathbf{w}^{gpt}$).
These results highlight that the data mixture predicted by \trex{} not only improves compression efficiency but also yields substantial computational savings during large-scale language model training.

\paragraph{Cost Efficiency of \trex{} in Finding the Optimal Data Mixture.}
A potential concern is whether \trex{}'s regression-based approach offers tokenizer training efficiency. Despite the initial cost of training 480 proxy tokenizers, \trex{} determines the optimal mixture in a single step, while AdaptMix~\cite{thakur2025art} requires about 20 large-scale iterations ($S{=}30$GB, $V{=}200$K) to converge. As shown in Appendix~\ref{appendix:cost_tokenizer}, this leads to a 52.2\% reduction in total training time, saving roughly 20 hours overall. These results demonstrate that \trex{} achieves significantly higher time efficiency with minimal computational overhead. Further implementation details are provided in Appendix~\ref{appendix:cost_tokenizer}.

\paragraph{Is \trex{} Effective in Domain-Specific Scenarios?}
In practice, it is important to train tokenizers that are effective in both multilingual and domain-specific environments~\cite{dagan2024getting, abagyan2025one}. We therefore examine whether the proposed \trex{} remains effective when the training is focused on a specific target domain. To this end, we used the \texttt{Pile} dataset with domain labels and trained the regression model to optimize compression performance on medical domain text. As shown in \autoref{tab:regression_results_in_medical}, the regression model of \trex{} achieved a Spearman rank correlation above 0.965 and a MAPE of 0.921. This indicates that \trex{} can accurately predict the compression behavior of data mixtures even within a specific domain.
Additional supporting experiments and detailed analyses of domain-specific performance are provided in Appendix~\ref{sec:appendix_regression_rank}.
\begin{table}[t]
\small
\centering
\begin{tabular}{llcc}
\toprule
\multicolumn{2}{c}{\textbf{Test Setting}} & \multirow{2}{*}{\textbf{Correlation} ($\rho$)} & \multirow{2}{*}{\textbf{MAPE}} \\
\cmidrule(lr){1-2} 
Data Size & Vocabulary & $\uparrow$ & $\downarrow$ \\
\midrule
1GB   & 64k    & 0.981   & 0.921  \\
% 5GB   & 64k    & 0.968   & -  \\
% 10GB  & 100k   & 0.970   & -  \\
% 30GB  & 200k   & 0.967   & -  \\
\bottomrule
\end{tabular}
\caption{Performance of the regression model in the medical domain.}
\label{tab:regression_results_in_medical}
\vspace{-0.2in}
\end{table}

\section{Conclusion}
\trex{} effectively addresses the fundamental challenge of identifying the optimal data mixture for tokenizer training. By leveraging a regression model trained on small-scale proxy tokenizers, it can accurately predict the compression performance of various data mixtures without the need for repeated full-scale training, thereby significantly reducing computational costs.
The regression model of \trex{} demonstrated high reliability, achieving a MAPE of less than 2\% and a Spearman rank correlation exceeding 0.97 when predicting tokenizer compression performance. Owing to this precise predictive capability, the tokenizer trained with the optimal mixture derived from \trex{} achieved up to a 12\% improvement in compression efficiency compared to heuristic approaches and data mixture used in tokenizers such as LLaMA3 and GPT-4o. This improvement was consistently observed across both in-distribution and out-of-distribution data.

\section*{Limitations}
\trex{} has a clear objective: to improve the compression performance of tokenizers. Better compression provides the practical benefit of reducing the total token count, thereby enhancing the training and inference speed of LLMs. However, compression performance is not always proportional to the resulting language model's downstream performance on tasks such as translation, summarization, or reasoning. This study does not examine how a tokenization method optimized solely for compression may affect a model's semantic understanding or complex reasoning capability. Investigating the relationship between compression efficiency and the overall model quality remains an important direction for future research.
Furthermore, while our experimental results (from 1GB/64k to 30GB/200k) strongly support the underlying assumption of rank invariance, further validation is required. It is necessary to determine if this assumption holds at more extreme scales (e.g., training data exceeding several hundred gigabytes, vocabularies larger than 500,000) or with entirely different sets of languages.
Finally, this research was conducted using a dataset composed of 19 languages. Although these languages span various families and writing systems, they do not represent the full linguistic diversity of the world. The relationship between data mixture and compression ratio could become more complex, particularly if a large number of morphologically rich languages (such as agglutinative or polysynthetic ones) or low-resource languages are included.

\section*{Acknowledgement}
This research was supported by the InnoCORE program of the Ministry of Science and ICT(N10250154) and the Top-Tier AI Global HRD invitation program (RS-2025-25461932) supervised by the IITP(Institute for Information \& Communications Technology Planning \& Evaluation)

\bibliography{anthology,custom}

@article{xie2024doremi,
  title={Doremi: Optimizing data mixtures speeds up language model pretraining},
  author={Xie, Sang Michael and Pham, Hieu and Dong, Xuanyi and Du, Nan and Liu, Hanxiao and Lu, Yifeng and Liang, Percy S and Le, Quoc V and Ma, Tengyu and Yu, Adams Wei},
  journal={Advances in Neural Information Processing Systems},
  volume={36},
  pages={69798--69818},
  year={2024}
}

@article{fan2023doge,
  title={Doge: Domain reweighting with generalization estimation},
  author={Fan, Simin and Pagliardini, Matteo and Jaggi, Martin},
  journal={arXiv preprint arXiv:2310.15393},
  year={2023}
}

@article{ye2024data,
  title={Data mixing laws: Optimizing data mixtures by predicting language modeling performance},
  author={Ye, Jiasheng and Liu, Peiju and Sun, Tianxiang and Zhan, Jun and Zhou, Yunhua and Qiu, Xipeng},
  journal={arXiv preprint arXiv:2403.16952},
  year={2024}
}

@inproceedings{liu2025regmix,
  author = {Liu, Qian and Zheng, Xiaosen and Muennighoff, Niklas and Zeng, Guangtao and Dou, Longxu and Pang, Tianyu and Jiang, Jing and Lin, Min},
  title = {RegMix: Data Mixture as Regression for Language Model Pre-training},
  booktitle = {International Conference on Learning Representations (ICLR)},
  year = {2025}
}

@inproceedings{
li2025data,
title={Data Mixing Optimization for Supervised Fine-Tuning of Large Language Models},
author={Yuan Li and Zhengzhong Liu and Eric P. Xing},
booktitle={Forty-second International Conference on Machine Learning},
year={2025},
url={https://openreview.net/forum?id=19kqoNoc2N}
}

@article{ke2017lightgbm,
  title={Lightgbm: A highly efficient gradient boosting decision tree},
  author={Ke, Guolin and Meng, Qi and Finley, Thomas and Wang, Taifeng and Chen, Wei and Ma, Weidong and Ye, Qiwei and Liu, Tie-Yan},
  journal={Advances in neural information processing systems},
  volume={30},
  year={2017}
}

@inproceedings{petrov2023fairness,
  author    = {Aleksandar Petrov and Emanuele La Malfa and Philip H.~S. Torr and Adel Bibi},
  title     = {Language Model Tokenizers Introduce Unfairness Between Languages},
  booktitle = {Advances in Neural Information Processing Systems 36 (NeurIPS 2023)},
  year      = {2023}
}

@article{stollenwerk2023fertility,
  author    = {Felix Stollenwerk},
  title     = {Training and Evaluation of a Multilingual Tokenizer for GPT-SW3},
  journal   = {CoRR},
  volume    = {abs/2304.14780},
  year      = {2023},
  url       = {https://arxiv.org/abs/2304.14780},
  eprinttype= {arXiv},
  eprint    = {2304.14780}
}

@article{scao2022bloom,
  author    = {Teven Le Scao and Angela Fan and Christopher Akiki and Ellie Pavlick and Suzana Ili\'{c} and Daniel Hesslow and Roman Castagn\'{e} and Alexandra Luccioni and others},
  title     = {BLOOM: A 176B-Parameter Open-Access Multilingual Language Model},
  journal   = {CoRR},
  volume    = {abs/2211.05100},
  year      = {2022},
  url       = {https://arxiv.org/abs/2211.05100},
  eprinttype= {arXiv},
  eprint    = {2211.05100}
}

@inproceedings{ali-etal-2024-tokenizer,
    title = "Tokenizer Choice For {LLM} Training: Negligible or Crucial?",
    author = {Ali, Mehdi  and
      Fromm, Michael  and
      Thellmann, Klaudia  and
      Rutmann, Richard  and
      L{\"u}bbering, Max  and
      Leveling, Johannes  and
      Klug, Katrin  and
      Ebert, Jan  and
      Doll, Niclas  and
      Buschhoff, Jasper  and
      Jain, Charvi  and
      Weber, Alexander  and
      Jurkschat, Lena  and
      Abdelwahab, Hammam  and
      John, Chelsea  and
      Ortiz Suarez, Pedro  and
      Ostendorff, Malte  and
      Weinbach, Samuel  and
      Sifa, Rafet  and
      Kesselheim, Stefan  and
      Flores-Herr, Nicolas},
    editor = "Duh, Kevin  and
      Gomez, Helena  and
      Bethard, Steven",
    booktitle = "Findings of the Association for Computational Linguistics: NAACL 2024",
    month = jun,
    year = "2024",
    address = "Mexico City, Mexico",
    publisher = "Association for Computational Linguistics",
    url = "https://aclanthology.org/2024.findings-naacl.247/",
    doi = "10.18653/v1/2024.findings-naacl.247",
    pages = "3907--3924"
}

@article{abagyan2025one,
  title={One Tokenizer To Rule Them All: Emergent Language Plasticity via Multilingual Tokenizers},
  author={Abagyan, Diana and Salamanca, Alejandro R and Cruz-Salinas, Andres Felipe and Cao, Kris and Lin, Hangyu and Locatelli, Acyr and Fadaee, Marzieh and {\"U}st{\"u}n, Ahmet and Hooker, Sara},
  journal={arXiv preprint arXiv:2506.10766},
  year={2025}
}

@article{reddy2025much,
  title={How much is enough? the diminishing returns of tokenization training data},
  author={Reddy, Varshini and Schmidt, Craig W and Pinter, Yuval and Tanner, Chris},
  journal={arXiv preprint arXiv:2502.20273},
  year={2025}
}

@article{goldman2024unpacking,
  title={Unpacking tokenization: Evaluating text compression and its correlation with model performance},
  author={Goldman, Omer and Caciularu, Avi and Eyal, Matan and Cao, Kris and Szpektor, Idan and Tsarfaty, Reut},
  journal={arXiv preprint arXiv:2403.06265},
  year={2024}
}

@article{thakur2025art,
  title={The Art of Breaking Words: Rethinking Multilingual Tokenizer Design},
  author={Thakur, Aamod and Nagpal, Ajay and Savarkar, Atharva and Pundalik, Kundeshwar and Dosi, Siddhesh and Sawarkar, Piyush and Thakur, Viraj and Saluja, Rohit and Desarkar, Maunendra Sankar and Ramakrishnan, Ganesh},
  journal={arXiv preprint arXiv:2508.06533},
  year={2025}
}

@inproceedings{dagan2024getting,
author = {Dagan, Gautier and Synnaeve, Gabriel and Rozi\`{e}re, Baptiste},
title = {Getting the most out of your tokenizer for pre-training and domain adaptation},
year = {2024},
publisher = {JMLR.org},
booktitle = {Proceedings of the 41st International Conference on Machine Learning},
articleno = {387},
numpages = {22},
location = {Vienna, Austria},
series = {ICML'24}
}

@article{seo2025does,
  title={How does a Language-Specific Tokenizer affect LLMs?},
  author={Seo, Jean and Kim, Jaeyoon and Byun, SungJoo and Shin, Hyopil},
  journal={arXiv preprint arXiv:2502.12560},
  year={2025}
}

@article{hayase2024data,
  title={Data mixture inference attack: BPE tokenizers reveal training data compositions},
  author={Hayase, Jonathan and Liu, Alisa and Choi, Yejin and Oh, Sewoong and Smith, Noah A},
  journal={Advances in Neural Information Processing Systems},
  volume={37},
  pages={8956--8983},
  year={2024}
}

@article{gowda2020finding,
  title={Finding the optimal vocabulary size for neural machine translation},
  author={Gowda, Thamme and May, Jonathan},
  journal={arXiv preprint arXiv:2004.02334},
  year={2020}
}

@article{ahia2023all,
  title={Do all languages cost the same? tokenization in the era of commercial language models},
  author={Ahia, Orevaoghene and Kumar, Sachin and Gonen, Hila and Kasai, Jungo and Mortensen, David R and Smith, Noah A and Tsvetkov, Yulia},
  journal={arXiv preprint arXiv:2305.13707},
  year={2023}
}

@article{bi2024deepseek,
  title={Deepseek llm: Scaling open-source language models with longtermism},
  author={Bi, Xiao and Chen, Deli and Chen, Guanting and Chen, Shanhuang and Dai, Damai and Deng, Chengqi and Ding, Honghui and Dong, Kai and Du, Qiushi and Fu, Zhe and others},
  journal={arXiv preprint arXiv:2401.02954},
  year={2024}
}

@article{lin2016dirichlet,
  title={On the dirichlet distribution},
  author={Lin, Jiayu},
  journal={Department of Mathematics and Statistics, Queens University},
  volume={40},
  year={2016}
}

@article{tsun2020probability,
  title={Probability and Statistics With Applications to Computing},
  author={Tsun, A},
  year={2020},
  publisher={University of Washington}
}

@article{messmer2025multilingdatacomp,
  title={Enhancing Multilingual LLM Pretraining with Model-Based Data Selection}, 
  author={Bettina Messmer and Vinko Sabolčec and Martin Jaggi},
  journal={arXiv},
  year={2025},
  url={https://arxiv.org/abs/2502.10361}, 
}

@misc{nllbteam2022languageleftbehindscaling,
      title={No Language Left Behind: Scaling Human-Centered Machine Translation}, 
      author={NLLB Team and Marta R. Costa-jussà and James Cross and Onur Çelebi and Maha Elbayad and Kenneth Heafield and Kevin Heffernan and Elahe Kalbassi and Janice Lam and Daniel Licht and Jean Maillard and Anna Sun and Skyler Wang and Guillaume Wenzek and Al Youngblood and Bapi Akula and Loic Barrault and Gabriel Mejia Gonzalez and Prangthip Hansanti and John Hoffman and Semarley Jarrett and Kaushik Ram Sadagopan and Dirk Rowe and Shannon Spruit and Chau Tran and Pierre Andrews and Necip Fazil Ayan and Shruti Bhosale and Sergey Edunov and Angela Fan and Cynthia Gao and Vedanuj Goswami and Francisco Guzmán and Philipp Koehn and Alexandre Mourachko and Christophe Ropers and Safiyyah Saleem and Holger Schwenk and Jeff Wang},
      year={2022},
      eprint={2207.04672},
      archivePrefix={arXiv},
      primaryClass={cs.CL},
      url={https://arxiv.org/abs/2207.04672}, 
}

@article{zuo2025falcon,
  title={Falcon-h1: A family of hybrid-head language models redefining efficiency and performance},
  author={Zuo, Jingwei and Velikanov, Maksim and Chahed, Ilyas and Belkada, Younes and Rhayem, Dhia Eddine and Kunsch, Guillaume and Hacid, Hakim and Yous, Hamza and Farhat, Brahim and Khadraoui, Ibrahim and others},
  journal={arXiv preprint arXiv:2507.22448},
  year={2025}
}

@inproceedings{
liu2025superbpe,
title={Super{BPE}: Space Travel for Language Models},
author={Alisa Liu and Jonathan Hayase and Valentin Hofmann and Sewoong Oh and Noah A. Smith and Yejin Choi},
booktitle={Second Conference on Language Modeling},
year={2025},
url={https://openreview.net/forum?id=lcDRvffeNP}
}

@misc{pundalik2025param1bharatgen29bmodel,
      title={PARAM-1 BharatGen 2.9B Model}, 
      author={Kundeshwar Pundalik and Piyush Sawarkar and Nihar Sahoo and Abhishek Shinde and Prateek Chanda and Vedant Goswami and Ajay Nagpal and Atul Singh and Viraj Thakur and Vijay Dewane and Aamod Thakur and Bhargav Patel and Smita Gautam and Bhagwan Panditi and Shyam Pawar and Madhav Kotcha and Suraj Racha and Saral Sureka and Pankaj Singh and Rishi Bal and Rohit Saluja and Ganesh Ramakrishnan},
      year={2025},
      eprint={2507.13390},
      archivePrefix={arXiv},
      primaryClass={cs.CL},
      url={https://arxiv.org/abs/2507.13390}, 
}

@misc{karthika2025multilingualtokenizationlensindian,
      title={Multilingual Tokenization through the Lens of Indian Languages: Challenges and Insights}, 
      author={N J Karthika and Maharaj Brahma and Rohit Saluja and Ganesh Ramakrishnan and Maunendra Sankar Desarkar},
      year={2025},
      eprint={2506.17789},
      archivePrefix={arXiv},
      primaryClass={cs.CL},
      url={https://arxiv.org/abs/2506.17789}, 
}

@article{olmo20242,
  title={2 OLMo 2 Furious},
  author={OLMo, Team and Walsh, Pete and Soldaini, Luca and Groeneveld, Dirk and Lo, Kyle and Arora, Shane and Bhagia, Akshita and Gu, Yuling and Huang, Shengyi and Jordan, Matt and others},
  journal={arXiv preprint arXiv:2501.00656},
  year={2024}
}

@article{hurst2024gpt,
  title={Gpt-4o system card},
  author={Hurst, Aaron and Lerer, Adam and Goucher, Adam P and Perelman, Adam and Ramesh, Aditya and Clark, Aidan and Ostrow, AJ and Welihinda, Akila and Hayes, Alan and Radford, Alec and others},
  journal={arXiv preprint arXiv:2410.21276},
  year={2024}
}

@article{grattafiori2024llama,
  title={The llama 3 herd of models},
  author={Grattafiori, Aaron and Dubey, Abhimanyu and Jauhri, Abhinav and Pandey, Abhinav and Kadian, Abhishek and Al-Dahle, Ahmad and Letman, Aiesha and Mathur, Akhil and Schelten, Alan and Vaughan, Alex and others},
  journal={arXiv preprint arXiv:2407.21783},
  year={2024}
}

@misc{kim2025kormokoreanopenreasoning,
      title={KORMo: Korean Open Reasoning Model for Everyone}, 
      author={Minjun Kim and Hyeonseok Lim and Hangyeol Yoo and Inho Won and Seungwoo Song and Minkyung Cho and Junhun Yuk and Changsu Choi and Dongjae Shin and Huije Lee and Hoyun Song and Alice Oh and Kyungtae Lim},
      year={2025},
      eprint={2510.09426},
      archivePrefix={arXiv},
      primaryClass={cs.CL},
      url={https://arxiv.org/abs/2510.09426}, 
}

@inproceedings{singh2025global,
  title={Global mmlu: Understanding and addressing cultural and linguistic biases in multilingual evaluation},
  author={Singh, Shivalika and Romanou, Angelika and Fourrier, Cl{\'e}mentine and Adelani, David Ifeoluwa and Ngui, Jian Gang and Vila-Suero, Daniel and Limkonchotiwat, Peerat and Marchisio, Kelly and Leong, Wei Qi and Susanto, Yosephine and others},
  booktitle={Proceedings of the 63rd Annual Meeting of the Association for Computational Linguistics (Volume 1: Long Papers)},
  pages={18761--18799},
  year={2025}
}

@inproceedings{lesci-etal-2025-causal,
address = {Vienna, Austria},
author = {Lesci, Pietro and Meister, Clara and Hofmann, Thomas and Vlachos, Andreas and Pimentel, Tiago},
booktitle = {Proceedings of the 63rd Annual Meeting of the Association for Computational Linguistics (Volume 1: Long Papers)},
doi = {10.18653/v1/2025.acl-long.1374},
editor = {Che, Wanxiang and Nabende, Joyce and Shutova, Ekaterina and Pilehvar, Mohammad Taher},
isbn = {979-8-89176-251-0},
month = {jul},
pages = {28325--28340},
publisher = {Association for Computational Linguistics},
title = {{Causal Estimation of Tokenisation Bias}},
url = {https://aclanthology.org/2025.acl-long.1374/},
year = {2025}
}

@article{siino-tinnirello-lacascia-2024-is-text,
author = {Siino, Marco and Tinnirello, Ilenia and {La Cascia}, Marco},
doi = {https://doi.org/10.1016/j.is.2023.102342},
issn = {0306-4379},
journal = {Information Systems},
number = {102342},
pages = {1--19},
title = {{Is text preprocessing still worth the time? A comparative survey on the influence of popular preprocessing methods on Transformers and traditional classifiers}},
url = {https://www.sciencedirect.com/science/article/pii/S0306437923001783},
volume = {121},
year = {2024}
}

@inproceedings{whittington-etal-2025-tokenisation,
address = {Vienna, Austria},
author = {Whittington, Philip and Bachmann, Gregor and Pimentel, Tiago},
booktitle = {Proceedings of the 63rd Annual Meeting of the Association for Computational Linguistics (Volume 1: Long Papers)},
doi = {10.18653/v1/2025.acl-long.1365},
editor = {Che, Wanxiang and Nabende, Joyce and Shutova, Ekaterina and Pilehvar, Mohammad Taher},
isbn = {979-8-89176-251-0},
month = {jul},
pages = {28133--28153},
publisher = {Association for Computational Linguistics},
title = {{Tokenisation is NP-Complete}},
url = {https://aclanthology.org/2025.acl-long.1365/},
year = {2025}
}
\bibliographystyle{acl_natbib}

\clearpage
\appendix

\section{Detailed Description of Datasets}\label{sec:appendix_datasets}
\begin{table}[h!]
\small
\centering
\begin{tabular}{llr}
\toprule
\textbf{Subset name} & \textbf{Language name} & \textbf{Disk size} \\
\midrule
rus\_Cyrl & Russian & 1.2T \\
cmn\_Hani & Chinese & 784G \\
deu\_Latn & German & 618G \\
spa\_Latn & Spanish & 515G \\
jpn\_Jpan & Japanese & 393G \\
fra\_Latn & French & 483G \\
ita\_Latn & Italian & 269G \\
por\_Latn & Portuguese & 222G \\
pol\_Latn & Polish & 168G \\
nld\_Latn & Dutch & 160G \\
ind\_Latn & Indonesian & 125G \\
tur\_Latn & Turkish & 100G \\
ces\_Latn & Czech & 104G \\
fas\_Arab & Persian & 69G \\
hun\_Latn & Hungarian & 79G \\
swe\_Latn & Swedish & 61G \\
ell\_Grek & Greek & 84G \\
dan\_Latn & Danish & 61G \\
vie\_Latn & Vietnamese & 59G \\
\bottomrule
\end{tabular}
\caption{Language Composition of the FineWeb2-HQ Dataset}
\label{tab:fineweb2-hq}
\end{table}
\paragraph{FineWeb2-HQ Dataset}
FineWeb2-HQ is a high-quality, model-filtered pretraining dataset designed for multilingual Large Language Models (LLMs). It is a subset of the FineWeb2 corpus, spanning 19 languages, and was created by selecting the top 10\% of documents in each language. The selection was based on scores from a deep learning classifier, which used XLM-RoBERTa embeddings to identify structured and knowledge-rich samples. The language composition of the dataset is detailed in \autoref{tab:fineweb2-hq}.

\paragraph{FLORES Dataset}
The FLORES (Facebook Low-Resource Translation Evaluation) dataset is a multilingual benchmark designed to evaluate machine translation quality between English and low-resource languages. FLORES contains translations of 3,001 sentences sourced from 842 distinct web articles, with each sentence averaging about 21 words.

\paragraph{Pile Dataset}
The Pile is an open-source dataset consisting of approximately 800 GB of diverse text designed for large-scale language model pretraining. It was curated to provide a high-quality, representative mixture of domains including academic papers (e.g., arXiv), web text (e.g., Wikipedia, StackExchange, HackerNews), code (e.g., GitHub), legal and medical documents, and more.
The dataset comprises 22 component corpora, each selected to balance domain diversity and textual quality, enabling models trained on The Pile to develop broad generalization and reasoning abilities across multiple knowledge domains. Due to copyright concerns, we utilize the 17 subsets that do not violate copyright issues.

\section{Compression Rate Prediction Result for the 1GB–64k Tokenizer}
\begin{figure}[h]
    \centering
    \includegraphics[width=\columnwidth]{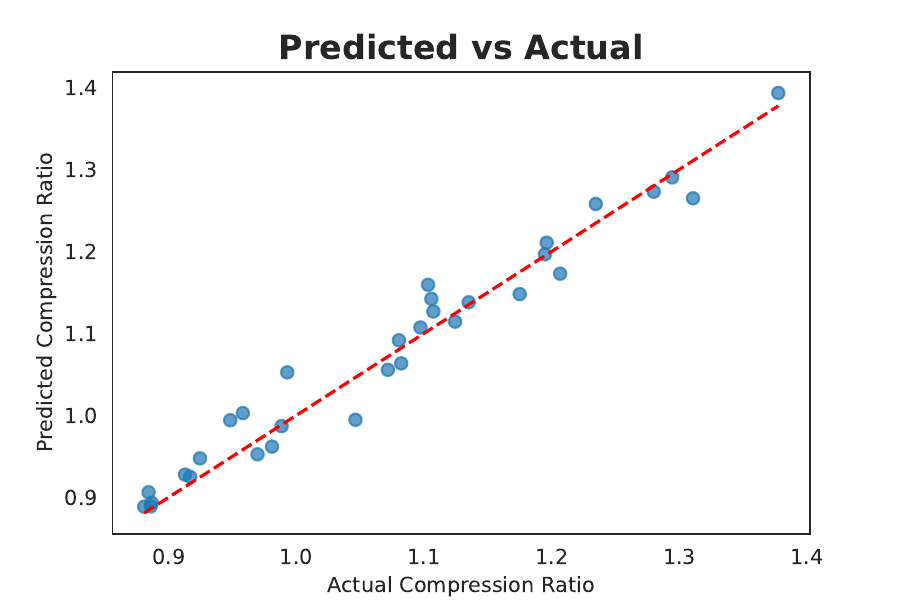}
    \caption{Compression rate prediction results for the 1GB–64k tokenizer using \trex{}}
    \label{fig:reg_1_64}
\end{figure}
Figure~\ref{fig:reg_1_64} illustrates the relationship between the actual compression rates and those predicted by the \trex{} regression model for the 1GB–64k tokenizer.  
The red dashed line denotes the ideal prediction line, and each blue dot represents the predicted and actual compression rate for a specific data mixture.  
Most points are closely aligned with the dashed line, indicating that the regression model accurately captures the real compression performance in this small-scale setting.  
This result confirms that \trex{} can effectively predict tokenizer compression behavior even with limited-scale training data.

\section{Rank Invariance}
\label{sec:appendix_rank_invariance}
To achieve higher compression performance in tokenizer training, it is essential to find the optimal data mixture $\mathbf{w}$. However, in a full-scale setting, exploring this requires repeatedly training tokenizers on various data mixtures, which demands enormous computational cost and time.
To effectively address the significant cost and time challenges, we propose rank invariance in tokenizer space as our core hypothesis. Rank invariance refers to the property that the performance ranking of tokenizers according to data mixture strategies $\mathbf{w}$ remains unchanged even when the training scale differs. For example, given two arbitrary data mixtures $\mathbf{w}_i$ and $\mathbf{w}_j$, if their compression performance ranking in small-scale settings $(S_s, V_s)$ is preserved in full-scale settings $(S_f, V_f)$, this relationship can be expressed as follows:
\begin{equation}
\begin{aligned}
    C_{\mathbf{w}_i}(S_s, V_s) > C_{\mathbf{w}_j}(S_s, V_s) \\
    \Longleftrightarrow \quad 
    C_{\mathbf{w}_i}(S_f, V_f) > &C_{\mathbf{w}_j}(S_f, V_f)
\end{aligned}
\label{eq:rank_invariance}
\end{equation}
To verify whether our proposed hypothesis holds in practice, we sampled 32 data mixtures across 19 languages and trained 32 tokenizers at various scales, ranging from $S=\text{1GB}, V=\text{64k}$ to $S=\text{30GB}, V=\text{200k}$. We then measured the compression performance of each trained tokenizer. To quantitatively assess how consistent the performance rankings are across different scale settings, we employed the Spearman Rank Correlation. Unlike metrics that focus on absolute performance differences, the Spearman Rank Correlation measures the correlation between rankings, making it the most suitable choice for numerically validating our core hypothesis of rank invariance.
\autoref{fig:spearman_corr_heatmap} presents the resulting heatmap. Across all setting pairs, the correlation coefficients were above 0.97, indicating strong consistency, even between the smallest and largest scales. This supports the Rank Invariance hypothesis and suggests that the optimal mixture in large-scale settings can be reliably predicted using only small-scale experiments.

\section{Evaluation of Regression Model in Large-Scale Settings}
\label{sec:appendix_regression_rank}
In the previous section, we confirmed that the proposed Rank Invariance hypothesis holds in actual tokenizer training.
A natural follow-up question is whether this invariance also extends to the relationship between the regression model's predicted compression and the actual compression scores.
To examine this, we conducted an experiment using a regression model trained at the proxy level—that is, under a small-scale configuration—to evaluate whether the rank correlation observed at small scales persists in real tokenizer performance.

\begin{table}[h]
\small
\centering
\begin{tabular}{llcc}
\toprule
\multicolumn{2}{c}{\textbf{Test Setting}} & \multirow{2}{*}{\textbf{Correlation} ($\rho$)} & \multirow{2}{*}{\textbf{MAPE}} \\
\cmidrule(lr){1-2} 
Data Size & Vocabulary & $\uparrow$ & $\downarrow$ \\
\midrule
\rowcolor{gray!15} 
\multicolumn{4}{c}{\textbf{Multilingual Domain}} \\
1GB   & 64k    &  0.979  & 1.989  \\
5GB   & 64k    &  0.970  & -  \\
10GB  & 100k   &  0.967  & -  \\
30GB  & 200k   &  0.960  & -  \\
\midrule
\rowcolor{gray!15} 
\multicolumn{4}{c}{\textbf{Medical Domain}} \\
1GB   & 64k    & 0.981   & 0.921  \\
5GB   & 64k    & 0.968   & -  \\
10GB  & 100k   & 0.970   & -  \\
30GB  & 200k   & 0.967   & -  \\
\bottomrule
\end{tabular}
\caption{Performance of the \trex{} regression model across different corpus and vocabulary scales under multilingual and medical-domain settings.
The model consistently exhibits high Spearman rank correlation ($\rho \geq 0.96$) between predicted and actual compression values, demonstrating that the Rank Invariance holds across both domains.}
\label{tab:regression_with_rank_invariance}
\end{table}

\autoref{tab:regression_with_rank_invariance} presents the results of regression models evaluated under both multilingual and medical-specific settings.
Across diverse scale configurations, the predicted compression scores from the regression model exhibited a consistently high rank correlation $\rho$ with the actual tokenizer compression results.
These findings indicate that the Rank Invariance—previously observed in real tokenizer training—also holds within the \trex{}'s regression model, reinforcing its validity as a scalable and reliable predictor of tokenizer performance.

\section{Tokenizer Compression Score in Medical Domain}\label{sec:appendix_regression_domain}
In Section~\ref{sec:analysis}, we further applied \trex{} to the medical domain to examine whether the optimal data mixture predicted by \trex{} leads to improved tokenizer performance.
\autoref{tab:comp_in_medical} presents the experimental results, showing that the tokenizer trained with the \trex{}-predicted mixture achieved a slight but consistent improvement in compression performance compared to existing baselines.
These findings demonstrate that \trex{} is effective not only in multilingual settings but also in domain-specific scenarios such as the medical domain.
\begin{table}[h]
\small
\centering
\begin{tabular}{lcc}
\toprule
& Baseline & \trex{} \\
\midrule
Compression & 0.911 & \textbf{0.904} \\
\bottomrule
\end{tabular}
\caption{Tokenizer compression in medical domain}
\label{tab:comp_in_medical}
\end{table}

\section{Cost Efficiency of \trex{} in Finding Optimal Data Mixture} \label{appendix:cost_tokenizer}
\label{app:cost-eff-token}
A potential concern regarding \trex{} is the initial overhead of building 480 proxy tokenizers to train its regression model. To address the question of its time cost-efficiency, we compared the total time required to obtain a final tokenizer with that of a recent iterative method, AdaptMix \cite{thakur2025art}, which requires approximately 20 iterations on the full-scale settings (S=30GB, V=200k).
\begin{figure}[h]
    \centering
    \includegraphics[width=\columnwidth]{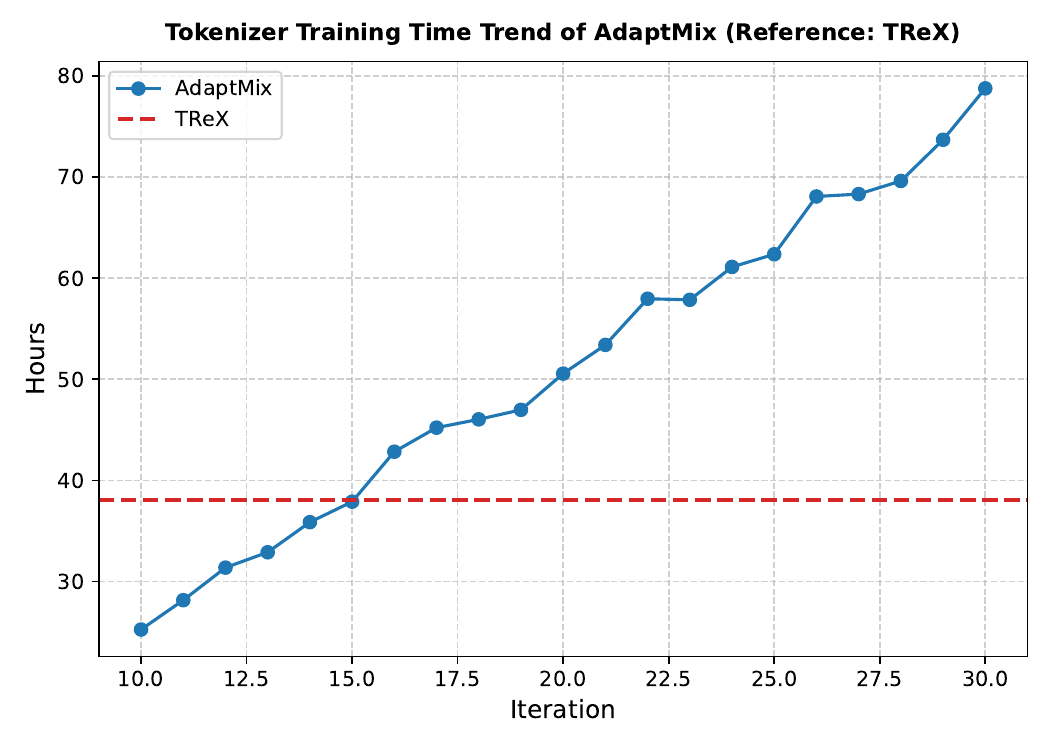}
    \caption{Time Consumption Comparison between AdaptMix and \trex{}}
    \label{fig:tokenizer_time_consumption}
\end{figure}

As illustrated in \autoref{fig:tokenizer_time_consumption}, our approach reduces the total training time by approximately 41 hours compared to AdaptMix.

Moreover, the advantage of \trex{} goes beyond mere time efficiency. Once trained, the regression model can rapidly simulate the performance of diverse data mixtures, enabling broad exploration of mixture ratios to identify globally optimal tokenizer configurations at negligible additional cost, a capability fundamentally distinct from iterative approaches that incur new costs for every trial.

\section{Cost Efficiency of \trex{} in Language Model Training}
As established in the previous section, an improved tokenization compression rate directly leads to higher efficiency in language model training. To quantify this effect, we designed an experiment to further illustrate the following point: a tokenizer with an improved compression encodes the same amount of raw text into fewer tokens, thereby reducing the total training FLOPs and, consequently, the overall training time.
We estimated this impact by first calculating a baseline — the total time required to train a model on 3 trillion (3T) tokens using a Uniform tokenizer on a cluster of 32 H100 GPUs.
\begin{figure}[h]
    \centering
    \includegraphics[width=\linewidth]{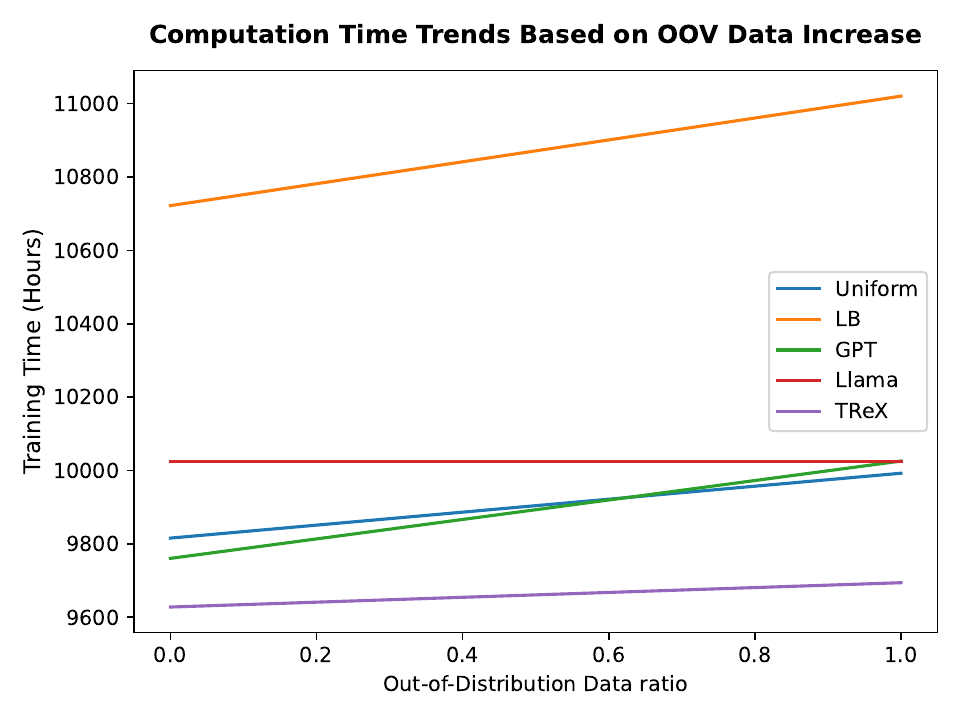}
    \caption{The x-axis represents the proportion of Out-of-Distribution (OOD) data within the pretraining corpus, and the y-axis denotes the corresponding training time. As shown, the tokenizer trained with \trex{} exhibits a slower increase in training time as the OOD ratio grows, compared to other tokenizers.}
    \label{fig:oov_time_consumption}
\end{figure}

\autoref{fig:oov_time_consumption} illustrates the projected time savings for various tokenizers based on their respective compression rates relative to this baseline. These results offer a concrete view of how tokenization compression impacts real-world training efficiency. Although the numerical differences in compression rates may appear small, the figure clearly shows that they yield substantial reductions in total training time, underscoring the significant efficiency gains achieved by the tokenizer optimized through \trex{}.

\section{Language Composition of Data Mixtures}
\label{sec:appendix_data_mixture_dist}
\begin{figure}
\centering
\includegraphics[width=\linewidth]{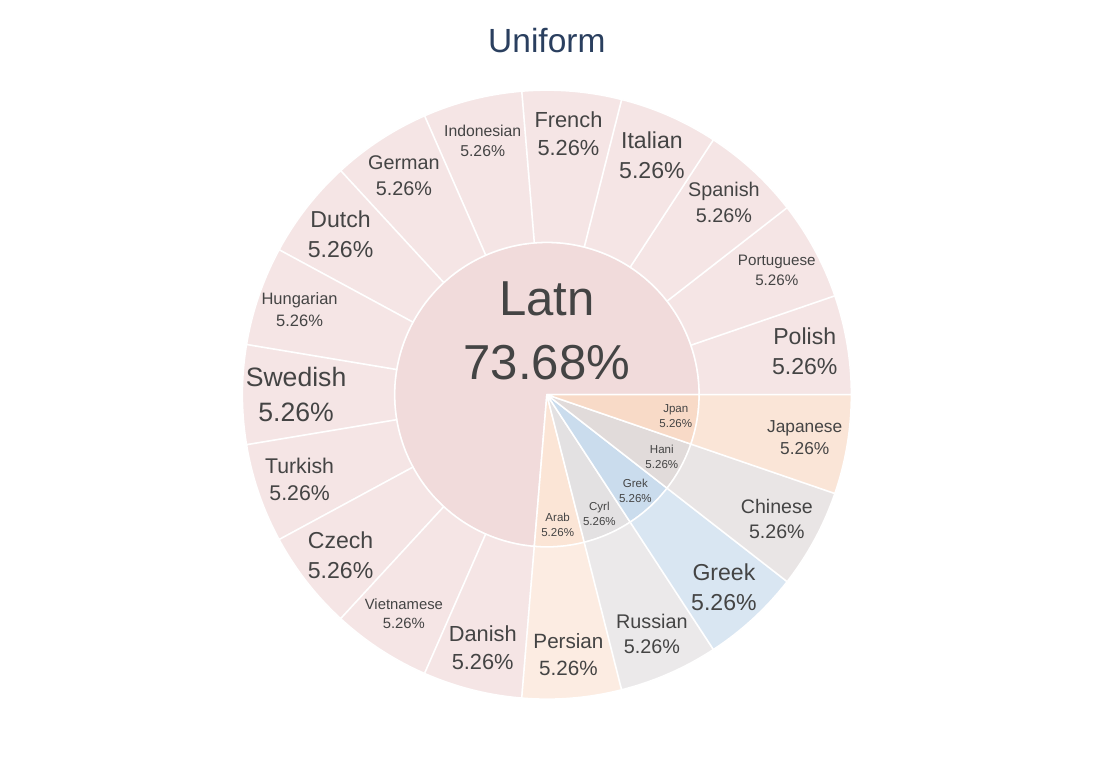}
\caption{Data mixture of Uniform}
\label{fig:uniform}
\end{figure}

\begin{figure}
    \centering
    \includegraphics[width=\linewidth]{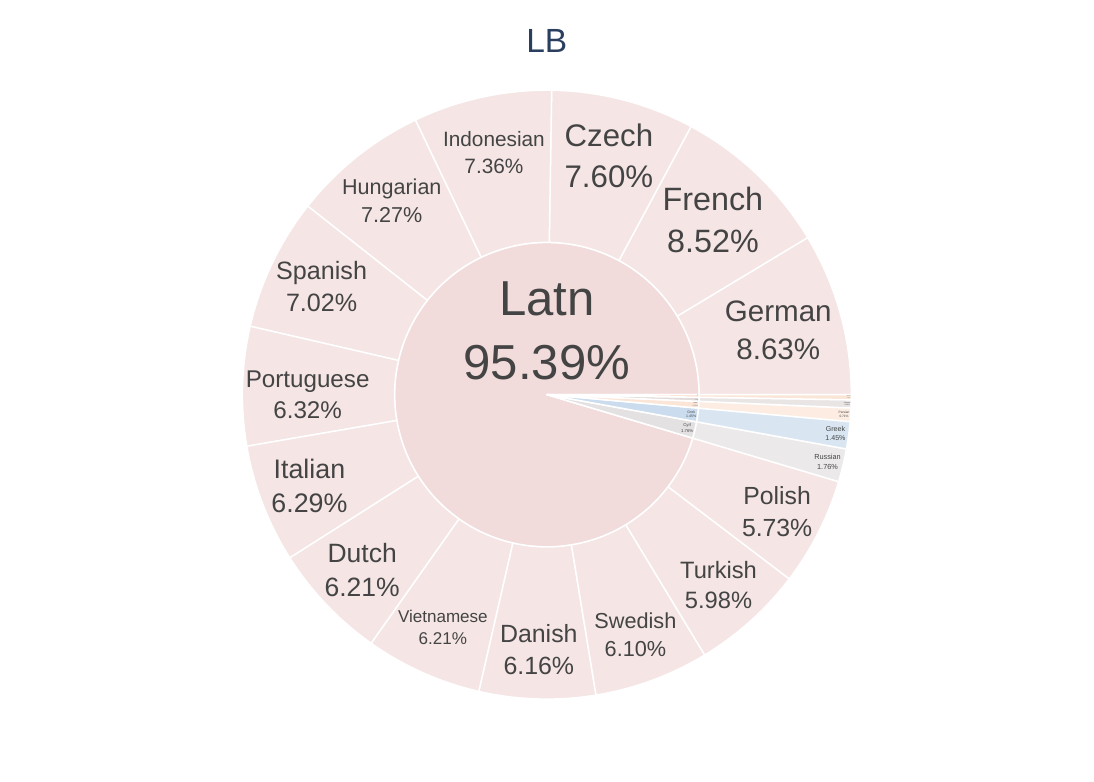}
    \caption{Data mixture of LB (entries below 1.7\% are omitted)}
    \label{fig:Lang_bucket}
\end{figure}

\begin{figure}
    \centering
    \includegraphics[width=\linewidth]{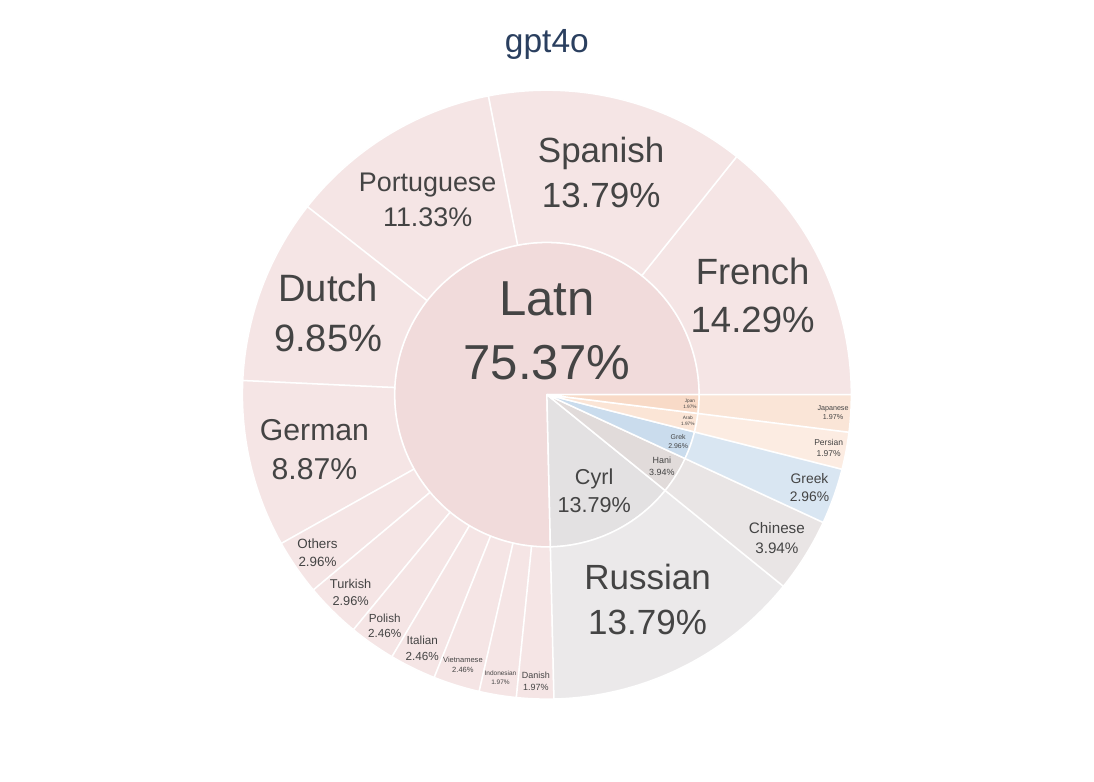}
    \caption{Data mixture of gpt-4o (entries below 1.7\% are omitted)}
    \label{fig:gpt4o}
\end{figure}

\begin{figure}
    \centering
    \includegraphics[width=\linewidth]{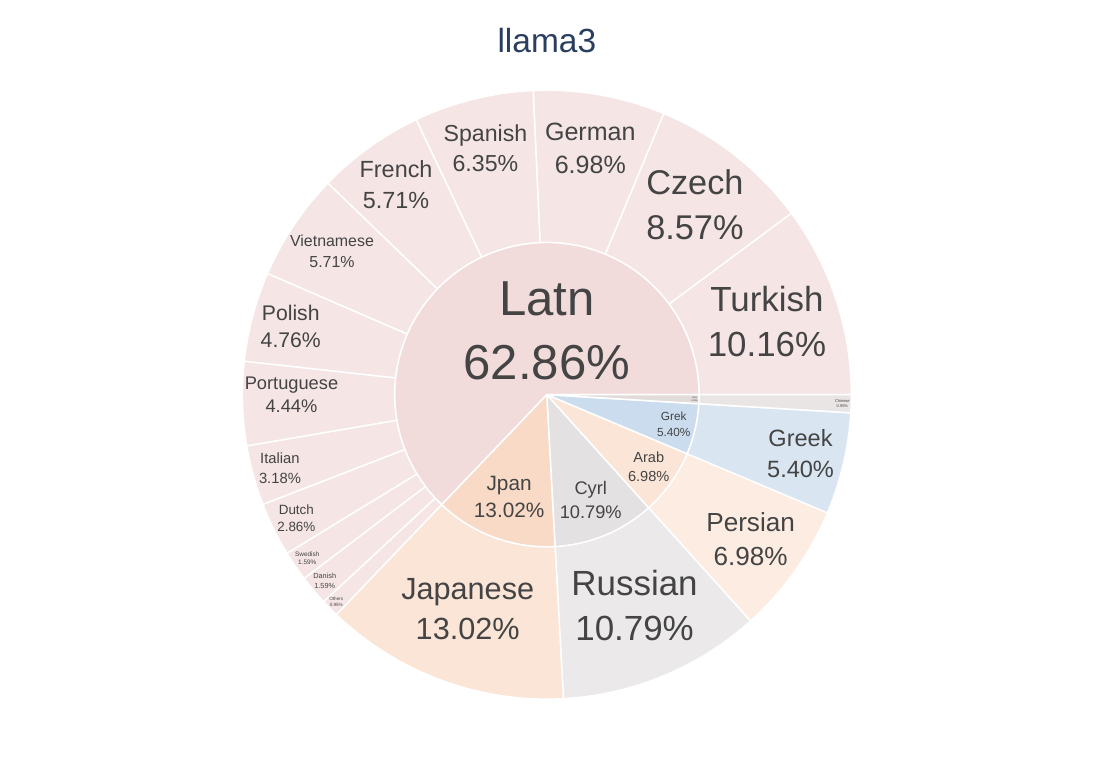}
    \caption{Data mixture of llama3 (entries below 1.7\% are omitted)}
    \label{fig:llama3}
\end{figure}

\begin{figure}
    \centering
    \includegraphics[width=\linewidth]{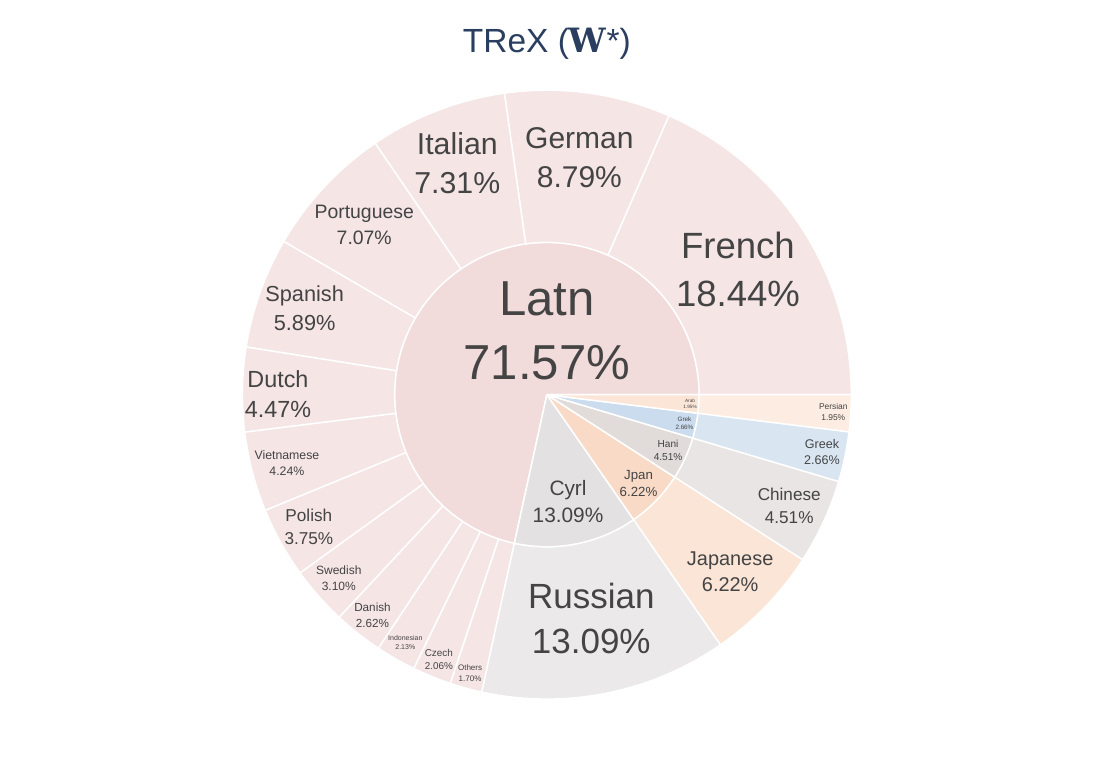}
    \caption{Data mixture of \trex{} (entries below 1.7\% are omitted)}
    \label{fig:trex}
\end{figure}

This section provides a visual analysis of the composition of each mixture used in Table~\ref{tab:comp-main-results}.
$\mathbf{w}^{LB}$, $\mathbf{w}^{gpt}$ and $\mathbf{w}^{llama}$ cover a wide range of languages, but in this study, we focus only on the languages included in the previously described dataset.
Figures~\ref{fig:uniform}, \ref{fig:Lang_bucket}, \ref{fig:gpt4o}, \ref{fig:llama3}, and \ref{fig:trex} show pie charts visualizing the language proportions of each mixture.

\paragraph{Correlation Between Language Distribution and Compression}

As summarized in \autoref{fig:language_distribution} and \autoref{tab:comp-main-results}, the five mixture models exhibit markedly different language distributions. $\mathbf{w}^{uni}$ allocates roughly equal proportions (~5.2\%) to all languages, whereas $\mathbf{w}^{LB}$ assigns greater weight to mid-frequency languages (e.g., DEU, FRA, NLD) while underrepresenting low-resource ones. $\mathbf{w}^{gpt}$ concentrates heavily on Latin-based languages such as French, Portuguese, and Spanish, while $\mathbf{w}^{llama}$ is biased towards non-Latin scripts (e.g., Japanese, Persian, Turkish).

\begin{table}[h]
\centering
\tiny
\begin{tabular}{llrrrrr}
\toprule
\textbf{\textsc{Lang}} & \textbf{\textsc{Char}} & \textbf{\textsc{$\mathbf{w}^{uni}$}} & \textbf{\textsc{$\mathbf{w}^{LB}$}} & \textbf{\textsc{$\mathbf{w}^{gpt}$}} & \textbf{\textsc{$\mathbf{w}^{llama}$}} & $\mathbf{w}^\ast$ \\
\midrule
\textbf{CES}  & Latn & 0.052 & 0.076 & 0.014  & 0.085   & 0.020 \\
\textbf{CMN}  & Hani & 0.052 & 0.004 & 0.039  & 0.009   & 0.045 \\
\textbf{DAN}  & Latn & 0.052 & 0.061 & 0.019  & 0.015   & 0.026 \\
\textbf{DEU}  & Latn & 0.052 & 0.086 & 0.088  & 0.069   & 0.087 \\
\textbf{ELL}  & Grek & 0.052 & 0.014 & 0.029  & 0.053   & 0.026 \\
\textbf{FAS}  & Arab & 0.052 & 0.007 & 0.019  & 0.069   & 0.019 \\
\textbf{FRA}  & Latn & 0.052 & 0.085 & 0.142  & 0.057   & 0.184 \\
\textbf{HUN}  & Latn & 0.052 & 0.072 & 0      & 0       & 0.003 \\
\textbf{IND}  & Latn & 0.052 & 0.073 & 0.019  & 0.009   & 0.021 \\
\textbf{ITA}  & Latn & 0.052 & 0.062 & 0.024  & 0.031   & 0.073 \\
\textbf{JPN}  & Jpan & 0.052 & 0.002 & 0.019  & 0.130   & 0.062 \\
\textbf{NLD}  & Latn & 0.052 & 0.062 & 0.098  & 0.028   & 0.044 \\
\textbf{POL}  & Latn & 0.052 & 0.057 & 0.024  & 0.047   & 0.037 \\
\textbf{POR}  & Latn & 0.052 & 0.063 & 0.113  & 0.044   & 0.070 \\
\textbf{RUS}  & Cyrl & 0.052 & 0.017 & 0.137  & 0.107   & 0.130 \\
\textbf{SPA}  & Latn & 0.052 & 0.070 & 0.137  & 0.063   & 0.058 \\
\textbf{SWE}  & Latn & 0.052 & 0.060 & 0.014  & 0.015   & 0.030 \\
\textbf{TUR}  & Latn & 0.052 & 0.059 & 0.029  & 0.101   & 0.013 \\
\textbf{VIE}  & Latn & 0.052 & 0.062 & 0.024  & 0.057   & 0.042 \\
\bottomrule
\end{tabular}
\caption{Overview of data mixtures used in our experiments. Each mixture defines a unique weighting configuration across multiple language corpora.}
\label{tab:appendix-data-mixtures}
\end{table}

In contrast, $\mathbf{w}^{\trex{}}$ maintains a balanced distribution between Latin and non-Latin language groups, while strategically increasing the presence of languages with lower domain redundancy (e.g., French, Russian, Portuguese). The resulting language distributions exhibit a statistically significant negative correlation with actual compression efficiency ($r=-0.47$, $p<0.05$). This implies that moderately increasing the proportion of certain languages tends to improve their compression performance (i.e., token length efficiency). Such results suggest that when a tokenizer sufficiently learns the segment redundancy of each language, subword segmentation becomes more stable.

\clearpage

\clearpage

\section{Mixture Diversity Analysis}
\label{sec:appendix-mixture-diversity}

\begin{figure}[ht]
\centering
\includegraphics[width=0.950\linewidth]{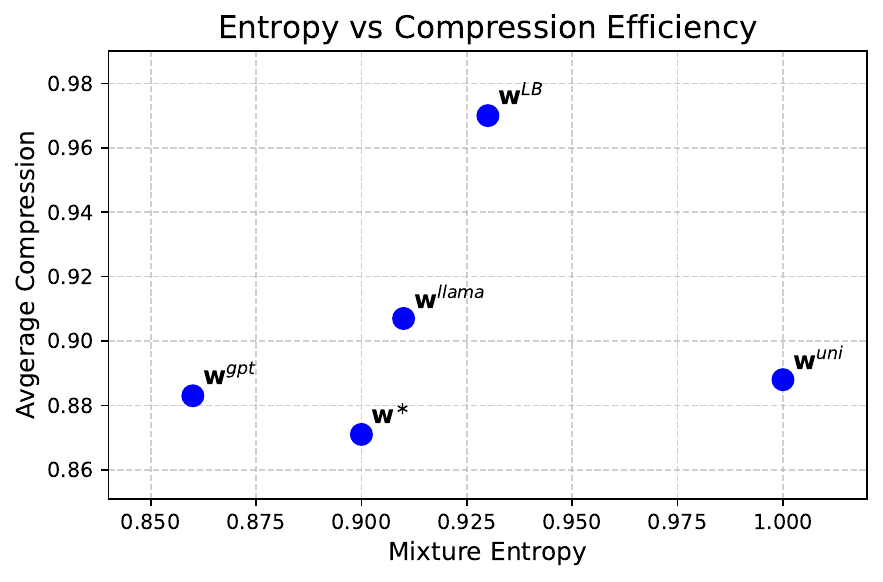}
\caption{
Relationship between mixture entropy (x-axis) and average compression ratio (y-axis) across five tokenizer configurations. 
A moderate level of entropy yields the most efficient tokenization, 
indicating that uniform distribution is not necessarily optimal for multilingual mixtures.
}
\label{fig:entropy_compression}
\end{figure}

To understand how language mixture diversity affects tokenizer efficiency, 
we analyze the statistical relationship between the entropy of each mixture distribution and its compression performance.
Figure~\ref{fig:entropy_compression} illustrates that mixtures with moderate entropy neither fully uniform nor overly skewed achieve the most efficient tokenization.  
This trend suggests that balanced yet biased allocation of languages improves subword segmentation across multilingual corpora.

\paragraph{Observation}
Uniform and LLaMA3 exhibit high mixture entropy (0.91--1.00), but yield only moderate compression efficiency (0.888--0.907).  
\trex{}, however, maintains moderate entropy (0.90) while achieving the best average compression ratio (0.871)  
and the lowest non-Latin subset average (0.814).  
This demonstrates that \trex{} learns an \textit{efficient bias} rather than relying on uniformity or hand-crafted heuristics.
\paragraph{Implication}
These results empirically validate that mixture optimization benefits from 
data-driven modeling of the non-linear relationship between language proportion and compression efficiency.  
In particular, the optimal point lies between complete uniformity and high skewness, 
highlighting the role of predictive mixture learning in multilingual tokenizer design.

\begin{table*}
\centering
\small
\begin{tabular}{lcccc}
\toprule
\textbf{Mixture Model} & \textbf{Entropy (↑)} & \textbf{Avg. Compression (↓)} & \textbf{Non-Latin Avg. (↓)} & $\Delta$ \textbf{vs. Uniform (↓)} \\
\midrule
\textsc{$\mathbf{w}^{uni}$}   & 1.00 & 0.888 & 0.848 & – \\
\textsc{$\mathbf{w}^{LB}$}    & 0.93 & 0.970 & 1.076 & +0.082 \\
\textsc{$\mathbf{w}^{gpt}$}   & 0.86 & 0.883 & 0.831 & -0.005 \\
\textsc{$\mathbf{w}^{llama}$} & 0.91 & 0.907 & 0.863 & +0.019 \\
\textsc{$\mathbf{w}^{\trex}$} & 0.90 & \textbf{0.871} & \textbf{0.814} & \textbf{-0.017} \\
\bottomrule
\end{tabular}
\caption{
Statistical summary of mixture entropy and compression performance across five tokenizer configurations. 
Entropy measures the diversity of language proportions within each mixture. 
\trex{} achieves the best overall and non-Latin compression efficiency while maintaining moderate entropy,
demonstrating that effective bias rather than uniformity leads to superior multilingual tokenization.
}
\label{tab:mixture_entropy}
\end{table*}

\subsection{Impact of tokenization and preprocessing on model behavior}
Recent work has shown that tokenization and text preprocessing play a far more active role in shaping language model behavior than previously assumed. \citet{lesci-etal-2025-causal} identify and quantify tokenization bias, showing that models trained with different vocabularies assign markedly different probabilities to the same character sequences. Framing this discrepancy as a causal effect, they estimate that the inclusion or exclusion of a single subword in a tokenizer's vocabulary can alter a model's predicted probability for its corresponding string by up to seventeenfold, revealing that tokenization is not a neutral preprocessing choice but a key determinant of model output. \citet{whittington-etal-2025-tokenisation} complement this empirical perspective with a formal analysis, demonstrating that the problem of optimal tokenization—whether defined in terms of vocabulary composition or merge sequence—is NP-complete. Their results explain why heuristic approaches such as byte-pair encoding and UnigramLM dominate in practice and highlight the inherent computational difficulty of designing universally optimal tokenizers. At a more applied level, \citet{siino-tinnirello-lacascia-2024-is-text} show that even seemingly minor preprocessing operations, including normalization, noise reduction, and punctuation handling, can substantially affect downstream performance, sometimes yielding differences exceeding 25\% in classification accuracy. Their findings reaffirm that preprocessing decisions remain critical in neural pipelines, affecting not only model robustness but also interpretability and computational efficiency. Taken together, these studies situate tokenization and preprocessing at the intersection of theory, practice, and linguistic representation, demonstrating that choices made at the input level fundamentally influence both the statistical and structural behavior of language models.

\section{Evaluation of Downstream Performance}
To address concerns regarding the correlation between tokenizer compression and downstream model performance, we conducted additional experiments to verify whether the increased compression efficiency of \trex{} affects the representative capabilities of Large Language Models.

Due to computational resource constraints, we performed evaluations under a budget-limited setting. We trained a 200M-parameter Transformer model from scratch using two different tokenizers: the standard LLaMA tokenizer (as a baseline) and the \trex{}-optimized tokenizer mixture.

To ensure a fair comparison of training volume, both models were trained on a fixed subset of the FineWeb2-HQ dataset, totaling 50 billion tokens as measured by the LLaMA tokenizer. This ensures that both models were exposed to the same amount of raw text data. For downstream evaluation, we utilized the Global MMLU~\citep{singh2025global} benchmark across 16 different languages to assess the model's cross-lingual reasoning and factual knowledge.

The results, summarized in \autoref{tab:downstream}, demonstrate that the model trained with the \trex{} tokenizer consistently achieves comparable or superior performance to the baseline across all tested languages. While the absolute performance gains are modest due to the restricted model size, the consistent trend suggests that \trex{}'s compression efficiency does not degrade—and in many cases enhances—the downstream utility of the language model.

\begin{table}[!h]
\small
\centering
\begin{tabular}{lccc}
\toprule
\textbf{Language} & $\textbf{w}^{llama}$ & $\textbf{w}^{\trex{}}$ & \textbf{Diff.} \\ \midrule
Russian (RUS)     & 22.95                    & \textbf{23.46}          & +0.51          \\
Polish (POL)      & 23.10                    & \textbf{23.75}          & +0.65          \\
Chinese (CMN)     & 22.91                    & \textbf{23.09}          & +0.18          \\
Dutch (NLD)       & \textbf{24.16}           & 23.69                   & -0.47          \\
German (DEU)      & 22.89                    & \textbf{24.17}          & +1.28          \\
Indonesian (IND)  & 22.95                    & \textbf{26.75}          & +3.80          \\
Japanese (JPN)    & 22.92                    & \textbf{22.95}          & +0.03          \\
Turkish (TUR)     & 22.97                    & \textbf{24.12}          & +1.15          \\
French (FRA)      & 23.31                    & \textbf{23.99}          & +0.68          \\
Czech (CES)       & 23.05                    & \textbf{23.17}          & +0.12          \\
Italian (ITA)     & 22.90                    & \textbf{23.56}          & +0.66          \\
Persian (FAS)     & \textbf{22.97}           & 22.93                   & -0.04          \\
Portuguese (POR)  & \textbf{22.97}           & 22.95                   & -0.02          \\
Swedish (SWE)     & 24.09                    & \textbf{24.51}          & +0.42          \\
Greek (ELL)       & 22.92                    & \textbf{23.01}          & +0.09          \\
Vietnamese (VIE)  & 24.72                    & \textbf{25.14}          & +0.42          \\ \midrule
\textbf{Average}  & 23.24                    & \textbf{23.83}          & \textbf{+0.59} \\ \bottomrule
\end{tabular}
\caption{Global MMLU performance of 200M-parameter models. Both models were trained on 50 billion tokens, with the dataset size determined based on the LLaMA tokenizer to ensure exposure to the same amount of raw text.}
\label{tab:downstream}
\end{table}

\end{document}